%% file: main.tex
\documentclass[lettersize,journal]{IEEEtran}

\ifCLASSOPTIONcompsoc
\usepackage[nocompress]{cite}
\else
\usepackage{cite}
\fi

\ifCLASSINFOpdf
\else
\fi

\usepackage[table]{xcolor}
\definecolor{Gray}{gray}{0.9}
\usepackage{gensymb}

\usepackage{amsmath,amssymb,amsfonts}
\usepackage{graphicx}
\usepackage{textcomp}
\usepackage{tikz}
\usetikzlibrary{bayesnet}

\usepackage{mathrsfs}
\usepackage[colorlinks,linkcolor=red,citecolor=blue]{hyperref}
\usepackage{amsmath}
\usepackage{amsfonts,subfigure}
\usepackage{amssymb}
\usepackage{multirow}
\usepackage{bm}
\usepackage{sansmath}
\usepackage{booktabs}
\usepackage[ruled]{algorithm2e}

\usepackage{graphicx}
\usepackage{amsthm}

\graphicspath{{figure/}}

\usepackage{bm}
\usepackage{color,soul}

\usepackage{tikz}

\usepackage{multirow}
\usepackage{adjustbox}
\usepackage{environ}
\usepackage{setspace}

\usepackage{enumitem}

\newcommand{\eqnref}[1]{Eq. (\ref{#1})}

\newcommand{\secref}[1]{Sec. \ref{#1}}
\newcommand{\tableref}[1]{Table~\ref{#1}}
  
\newcommand{\figref}[1]{Figure~\ref{#1}} 
\graphicspath{{image/}}
\DeclareMathOperator{\flops}{FLOPs}
\usepackage{makecell}
\definecolor{Gray}{gray}{0.9}
\usepackage{graphicx}

\begin{document}
	
	\title{STGformer: Efficient Spatiotemporal Graph Transformer for Traffic Forecasting }
	
	%
	%
	%
	%
	

	\author{Hongjun~Wang,
	Jiyuan~Chen,
	Tong~Pan,
	Zheng~Dong,\\
	Lingyu~Zhang,
	Renhe~Jiang,
	and
	Xuan Song
	\IEEEcompsocitemizethanks{
		\IEEEcompsocthanksitem Hongjun Wang, Jiyuan Chen,  Tong Pan,  Zheng Dong  and Lingyu Zhang are with (1) SUSTech-UTokyo Joint Research Center on Super Smart City, Department of Computer Science and Engineering
		(2) Research Institute of Trustworthy Autonomous Systems, Southern University of Science and Technology (SUSTech), Shenzhen, China.
		E-mail: {wanghj2020,11811810}@mail.sustech.edu.cn, pant@sustech.edu.cn,  zhengdong00@outlook.com, and zhanglingyu@didiglobal.com. 
		\IEEEcompsocthanksitem Xuan Song is with (1) School of Artificial Intelligence, Jilin University  (2) Research Institute of Trustworthy Autonomous Systems, Southern University of Science and Technology (SUSTech), Shenzhen, China. Email: songxuan@jlu.edu.cn.
		\IEEEcompsocthanksitem R. Jiang is with Center for Spatial Information Science, University of
		Tokyo, Tokyo, Japan. Email: jiangrh@csis.u-tokyo.ac.jp.
		\IEEEcompsocthanksitem Corresponding to  Xuan Song;
	}
}

	%
	%

\markboth{Journal of \LaTeX\ Class Files,~Vol.~XX, No.~X, August~201X}%
{Shell \MakeLowercase{\textit{et al.}}: Bare Demo of IEEEtran.cls for Computer Society Journals}
%

\maketitle
\begin{abstract}
Traffic forecasting is a cornerstone of smart city management, enabling efficient resource allocation and transportation planning. Deep learning, with its ability to capture complex nonlinear patterns in spatiotemporal (ST) data, has emerged as a powerful tool for traffic forecasting. While graph neural networks (GCNs) and transformer-based models have shown promise, their computational demands often hinder their application to real-world road networks, particularly those with large-scale spatiotemporal interactions.
To address these challenges, we propose a novel spatiotemporal graph transformer (STGformer) architecture. STGformer effectively balances the strengths of GCNs and Transformers, enabling efficient modeling of both global and local traffic patterns while maintaining a manageable computational footprint.  Unlike traditional approaches that require multiple attention layers, STG attention block captures high-order spatiotemporal interactions in a single layer, significantly reducing computational cost.
In particular, STGformer achieves a 100x speedup and a 99.8\% reduction in GPU memory usage compared to STAEformer during batch inference on a California road graph with 8,600 sensors. We evaluate STGformer on the LargeST benchmark and demonstrate its superiority over state-of-the-art Transformer-based methods such as PDFormer and STAEformer, which underline STGformer's potential to revolutionize traffic forecasting by overcoming the computational and memory limitations of existing approaches, making it a promising foundation for future spatiotemporal modeling tasks. \textcolor{magenta}{Codes are available at \textcolor{black}{\href{https://github.com/Dreamzz5/STGformer}{GitHub}}}
\end{abstract}
	

\section{Introduction}
\label{sec:intr}
\IEEEPARstart{S}{patiotemporal} graph neural networks \cite{li2017diffusion,wu2019graph,yan2018spatial,jain2016structural} have shown exceptional potential and have become a preferred method for making precise traffic predictions by leveraging graph neural networks to capture spatial relationships between sensors and utilize sequential models to learn temporal patterns. Recently, the emergence of Transformer-based architectures~\cite{liu2023spatio,jiang2023pdformer,xu2020spatial} greatly challenges the dominance of GCNs and become state of the arts architecture in traffic forecasting.
However, recent work  \cite{liu2024largest} have prove that most of ST-model, including Transformer, have failed when meeting large scale graph because current traffic datasets used for benchmarking are relatively small when compared to the complexity and scale of actual traffic networks. 
For instance, popular benchmark datasets like PEMS series \cite{guo2019attention}, MeTR-LA, PEMS-Bay \cite{li2017diffusion} consist of only a few hundred nodes and edges. In contrast, real-world traffic systems, such as  Caltrans Performance Measurement System in California, USA \cite{chen2001freeway}, incorporate nearly 20,000 active sensors.  Consequently, as traffic forecasting models are predominantly developed using these limited datasets, They often do not consider the computational overhead of the model 
fail to scale up to larger sensor networks, presenting a significant challenge in the field.  
\begin{figure}[t]
	\centering
	\includegraphics[width=1\linewidth]{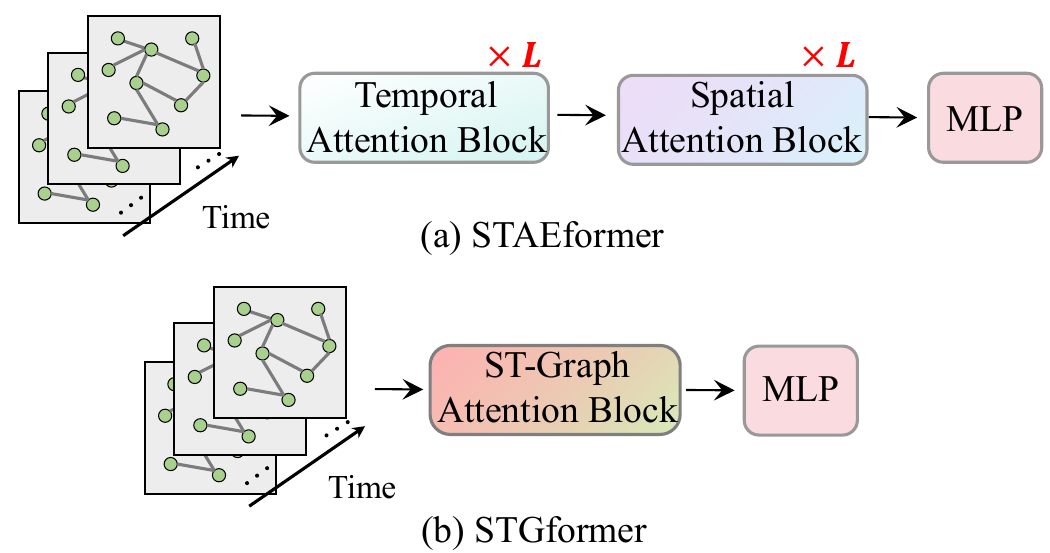}
	\caption{Unlike STAEformer, which relies on stacked $2L$ layers, STGformer integrates graph and spatiotemporal attention mechanisms, achieving superior performance with only \textbf{ a single layer}.}
	\label{fig:motivation}
\end{figure}

\begin{figure*}[t]
	\centering
	\includegraphics[width=0.85\linewidth]{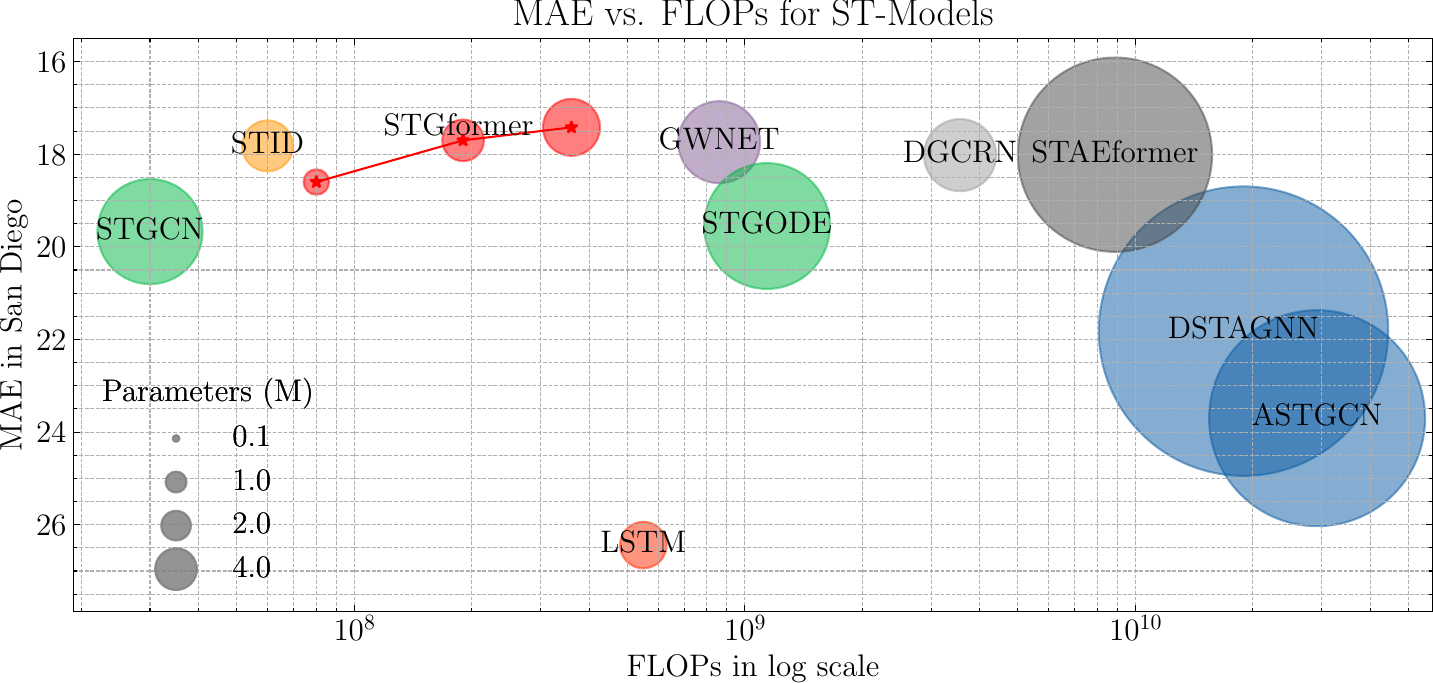}
	\caption{\textbf{MAE vs. FLOPs on traffic flow prediction in San Diego dataset \cite{liu2024largest}.} The bubble size indicate the model parameter number. We can observe that the standard graph convolution operation based method like ST-GCN achieve high computation efficiency but with inferior performance. And, the Transformer-based method achieve SOTA but bring large additional computation cost than GCN-basd method. }
	\label{fig:sd_performance}
\end{figure*}

As depicted in Figure \ref{fig:sd_performance}, we conducted a comparative analysis of leading methods on the San Diego dataset within the LargeST framework \cite{liu2024largest}. Our evaluation included performance, model parameters, and computational costs. While Transformer-based models, such as STAEformer \cite{liu2023spatio}, demonstrated superior performance, they exhibited significantly higher computational demands compared to GCN-based approaches like STGCN \cite{yu2018spatio}.
Based on these findings, we posit that the success of GCN and Transformer methods can be attributed to their respective strengths in capturing global spatial interactions and input-adaptive long-range dependencies \cite{yang2021focal}. Standard graph convolution operations \cite{hammond2011wavelets} effectively aggregate local neighbor features with high orders interaction \cite{yu2018spatio}. However, their reliance on local information limits their ability to consider global context in traffic scenarios, potentially hindering performance. In contrast, self-attention mechanisms, while disregarding explicit graph structures \cite{liu2023spatio}, implicitly capture global spatial interactions through successive matrix multiplications. This, however, introduces substantial computational overhead, making it challenging to scale to large-scale real-world traffic networks.

Previously, the overhead associated with global attention mechanisms remained within an acceptable range due to the use of only small-scale datasets for validation. However, the recent introduction of LargeST has exacerbated this issue. The global attention mechanism typically exhibits quadratic time and space complexity as the number of nodes increases, while the computational graph experiences exponential growth with an increasing number of layers. A potential compromise is to employ advanced techniques to partition interconnected nodes into smaller mini-batches, thereby reducing computational overhead \cite{nodeformer, wu2023difformer, survey-graphTransformer, chen2023nagphormer}. Nonetheless, this strategy results in longer training times due to the smaller mini-batches.
In this paper, we propose a more efficient model, which demonstrates exceptional competitiveness across seven traffic benchmarks, ranging from small-scale PEMS \cite{guo2019attention} datasets to large-scale datasets like LargeST \cite{liu2024largest}, utilizing only a single layer with linear spatiotemporal global attention. Despite the model's simplicity, it retains the full expressive capability necessary to capture all interactions among graph convolutions and attention mechanisms. Furthermore, our findings indicate that using fewer parameters can enhance generalization capabilities. Our research demonstrates that STGformer maintains strong performance even when tested on data from LargeST one year later.

In \figref{fig:motivation}, we specifically examine the differences between STGformer and STAEformer. Firstly, STAEformer predominantly relies on stacked \(2L\) layers and utilizes a spatiotemporal separable attention mechanism to achieve higher-order interactions by deepening the model. As previously noted, this approach incurs substantial computational overhead. In contrast, we introduce a more efficient attention module that integrates both graph and spatiotemporal attention mechanisms. Specifically, we conceptualize the temporal and spatial dimensions as a unified entity, employing the same query, key, and value in the attention mechanism to facilitate efficient spatiotemporal attention computation, which markedly reduces computational overhead compared to the separate treatment of these dimensions.
STGformer surpasses current state-of-the-art models by leveraging graph information, allowing for efficient computation using only a single layer of the attention module. Furthermore, we adopt linear attention \cite{katharopoulos2020Transformers, wang2020linformer, wu2024simplifying}, which replaces the softmax operation of the standard attention mechanism with decomposed inner products, thereby reducing the computational complexity from quadratic to linear. This significantly alleviates memory consumption and computational burden when handling large-scale spatiotemporal datasets.
The main contributions of this paper are summarized as follows:

\begin{itemize}
	\item[$\bullet$]  We propose a novel STG-attention  that efficiently captures high-order spatiotemporal interactions for both global and local patterns in a single layer, unlike previous methods requiring multiple stacked layers.
	
	\item[$\bullet$] STGformer combines the advantages of GCNs and Transformers while maintaining low computational and memory costs, significantly improving efficiency in processing large-scale traffic graphs compared to existing methods.
	
	\item[$\bullet$] STGformer outperforms state-of-the-art Transformer-based methods on the LargeST benchmark, demonstrating remarkable efficiency by being 100$\times$ faster and using 99.8\% less GPU memory than STAEformer during batch inference on California's traffic network.
\end{itemize}


\begin{figure*}[t]
	\centering
	\includegraphics[width=1\linewidth]{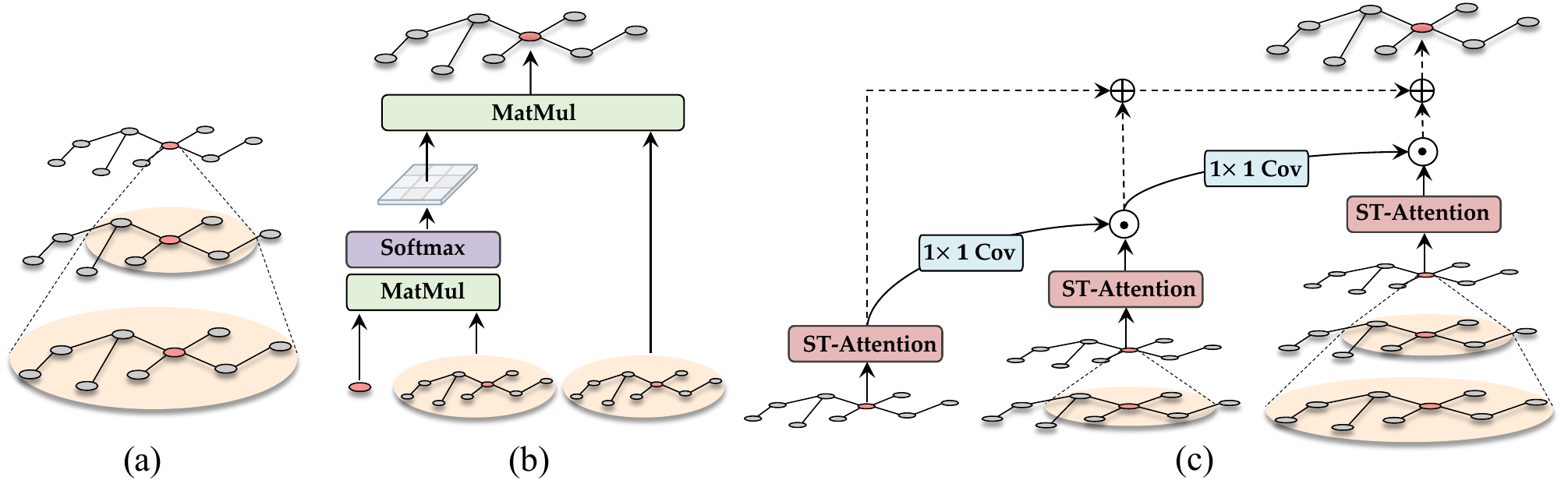}
	\caption{\textbf{Illustration the main idea of STGformer.} We illustrate typical spatial modeling in graphs, highlighting various levels of interactions and receptive fields.  (a) Conventional Graph Convolution~\cite{yu2018spatio,guo2019attention} explicitly handles arbitrary-order spatial interaction but within a limited receptive area. (b) Spatiotemporal Transformer ~\cite{liu2023spatio,jiang2023pdformer} performs interactions up to two orders through two consecutive matrix multiplications, covering a broad receptive area. (c) Our STGformer achieves high-order global spatial interactions with structure information  by integrating graph convolutional networks with Transformer architecture.}
	\label{fig:fig1}
\end{figure*}

\section{Related Work}

\subsection{Traffic Forecasting}
Deep neural networks have become the predominant approach for traffic forecasting \cite{li2018diffusion,cao2020spectral,liu2022contrastive,liu2023we,fang2023spatio}, typically combining graph neural networks (GNNs) with either Recurrent Neural Networks (RNNs) or Temporal Convolutional Networks (TCNs) to capture complex spatio-temporal patterns in traffic data.
RNN-based models, such as DCRNN \cite{li2018diffusion}, incorporate diffusion convolution with GRU layers to model spatial and temporal dependencies. Extensions of this approach include ST-MetaNet \cite{pan2019urban}, which employs meta-learning with graph attention networks, and AGCRN \cite{bai2020adaptive}, which introduces node-specific adaptive parameters in graph convolution.
To improve computational efficiency, TCN-based models like STGCN \cite{yu2018spatio} and GWNet \cite{wu2019graph} have adopted dilated causal convolutions for temporal modeling. These architectures demonstrate faster training times and competitive performance on various benchmarks.
Attention mechanisms have been integrated into models such as ASTGCN \cite{guo2019attention} and STAEformer \cite{liu2023spatio} to better capture long-range dependencies and complex spatio-temporal interactions. These approaches have shown improved performance in handling diverse traffic patterns.
Recent research directions include the integration of GNNs with neural ordinary differential equations for continuous modeling of spatio-temporal dependencies \cite{fang2021spatial,choi2022graph}, and the development of dynamic adjacency matrices to reflect changing relationships over time \cite{li2021dynamic,shao2022decoupled}. These emerging approaches aim to address limitations of previous models and enhance the adaptability and interpretability of traffic forecasting systems.

\subsection{Graph Transformer}
Graph Transformers have emerged as a powerful class of models for learning on graph-structured data, with several surveys reviewing different aspects of these models. The incorporation of graph structure into Transformer architectures has been explored through various graph inductive biases, as discussed by Dwivedi et al. \cite{dwivediGeneralizationTransformerNetworks2020a} and Rampášek et al. \cite{rampasekRecipeGeneralPowerful2022a}, who provided comprehensive overviews of node positional encodings, edge structural encodings, and attention bias. In terms of graph attention mechanisms, Velickovic et al. \cite{velickovicGraphAttentionNetworks2017c} introduced graph attention networks (GAT) leveraging multi-head attention for node classification, while subsequent works like GATv2 \cite{brodyHowAttentiveAre2021a} addressed limitations in GAT's expressiveness. The literature has explored various types of graph Transformers, including shallow models like GAT and GTN \cite{yunGraphTransformerNetworks2019d}, deep architectures stacking multiple attention layers \cite{gongHierarchicalGraphTransformerbased2020,yangTransformersOptimizationPerspective2022}, scalable versions addressing efficiency challenges for large graphs \cite{rampasekRecipeGeneralPowerful2022a,wuNodeformerScalableGraph2022a}, and pre-trained models leveraging self-supervised learning on large graph datasets \cite{liGraphixt5MixingPretrained2023,rongSelfsupervisedGraphTransformer2020a}. Graph Transformers have demonstrated promising results across various domains, including protein structure prediction in bioinformatics \cite{guHierarchicalGraphTransformer2023,pepeUsingGraphTransformer2023}, entity resolution in data management \cite{yaoEntityResolutionHierarchical2022b,yingTransformersReallyPerform2021i}, and anomaly detection in temporal data \cite{liuAnomalyDetectionDynamic2021a,xuAnomalyTransformerTime2022}. Despite their success, recent surveys \cite{mullerAttendingGraphTransformers2023e,rampasekRecipeGeneralPowerful2022a} have highlighted ongoing challenges in scalability, generalization, interpretability, and handling dynamic graphs, indicating that addressing these issues remains an active area of research in the graph learning community. At the same time, the Transformer based on spatiotemporal graph correlation has not yet been explored in the field of traffic prediction.

\section{Preliminaries} 
\subsection{Problem Statement}
In this paper, we formalize the representation of a graph as $\mathcal{G}=(\mathcal{V},\mathcal{E}, A)$, where $\mathcal{V}$ denotes the set of nodes with cardinality $N=|\mathcal{V}|$, $\mathcal{E} \subseteq \mathcal{V}\times \mathcal{V}$ defines the set of edges, and $A \in \mathbb{R}^{N \times N}$ represents the adjacency matrix. The dynamic nature of the graph is captured by a time-dependent feature matrix $\mathbf{X}_t\in \mathbb{R}^{|\mathcal{V}| \times \mathcal{C}}$ at each discrete time step $t$, where $\mathcal{C}$ denotes the dimensionality of node features (e.g., traffic flow, vehicle speed, and road occupation).
Traffic forecasting can be formally expressed as a function:
\begin{align}
	f: \left[\mathbf{X}_{(t-T): t}, \mathcal{G}\right] \mapsto \mathbf{X}_{(t+1):(t+S)},
\end{align}
where $T$ and $S$ represent the input and output sequence lengths, respectively, which encapsulates the temporal dependency by considering a historical window of $T$ time steps and predicting future states for $S$ time steps ahead.

\subsection{Spatiotemporal Graph Convolution} Graph neural networks (GNNs) have the advantage of aggregating node neighborhood contexts to generate spatial representations, which is earliest method introduced to spatiotemporal graph modeling \cite{yan2018spatial,yu2018spatio,li2018diffusion}. Formally, let $*_{\mathcal{G}}$ is graph convolution operator \cite{hammond2011wavelets}, which is reformulated as
\begin{align}\label{eqn:gcn}
	\Theta *_{\mathcal{G}} h = \Theta(\mathcal{L}) x \approx \sum_{k=0}^{K-1} \theta_k T_k(\tilde{\mathcal{L}}) h,  
\end{align}
where $\theta = [\theta_0, \ldots, \theta_{K-1}] \in \mathbb{R}^K$ is a vector of polynomial coefficients
$K$ denotes the kernel size of the graph convolution
$T_k(\tilde{\mathcal{L}}) \in \mathbb{R}^{N \times N}$ represents the $k$-th order Chebyshev polynomial evaluated at the rescaled Laplacian $\tilde{\mathcal{L}} = 2\mathcal{L} / \lambda_{\max} - I_N$
$\mathcal{L} = D^{-1/2}(D-A)D^{-1/2}$ is the normalized graph Laplacian
$D$ is the degree matrix of the graph
$\lambda_{\max}$ is the largest eigenvalue of $\mathcal{L}$
$I_N$ is the $N \times N$ identity matrix. This approach enables the efficient computation of $K$-localized convolutions by leveraging polynomial approximation, effectively capturing the local structure of the graph within a $K$-hop neighborhood.

\subsection{Spatiotemporal Self-Attention Layer}\label{sec:stae} Instead of GCNs \cite{yu2018spatio,wu2019graph,guo2019attention}, which aggregate neighboring features using static convolution kernels, Transformers \cite{jiang2023pdformer,liu2023spatio} employ multi-head self-attention to dynamically generate weights that mix spatial and temporal signals. 
Formally, given a hidden spatiotemporal representation $h \in \mathbb{R}^{T \times N \times C}$, where $T$ is the number of time frames, $N = |\mathcal{V}|$ is the number of graph nodes, and $C$ denotes the channel dimension, we can formulate the spatiotemporal self-attention mechanism as follows:
For spatial self-attention, the query, key, and value matrices are derived as: $Q_s = h W_Q^s$, $K_s = h W_K^s$, and $V_s = h W_V^s$, where $W_Q^s, W_K^s, W_V^s \in \mathbb{R}^{C \times C}$ are learnable parameter matrices. The spatial self-attention scores are then computed as: $A^s = \text{Softmax}(Q_s K_s^T / \sqrt{C})$, where $A^s \in \mathbb{R}^{T \times N \times N}$ captures spatial dependencies across nodes.
Similarly, for temporal self-attention, we have: $Q_t = h W_Q^t$, $K_t = h W_K^t$, and $V_t = h W_V^t$, with $W_Q^t, W_K^t, W_V^t \in \mathbb{R}^{C \times C}$ being distinct learnable parameters. The temporal self-attention scores are calculated as: $A^t = \text{Softmax}(Q_t^T K_t / \sqrt{C})$, where $A^t \in \mathbb{R}^{N \times T \times T}$ captures temporal dependencies across different time horizons.
This formulation allows for the dynamic generation of attention weights that simultaneously consider both spatial and temporal contexts, enabling the model to adapt to varying spatiotemporal patterns in the input data.

\subsection{Analysis of GCN and Transformer Flops}
In analyzing the computational complexity of spatiotemporal graph modeling techniques, we observe distinct characteristics between graph convolution and self-attention mechanisms. The spatiotemporal graph convolution, utilizing Chebyshev polynomials, exhibits a computational complexity of $\mathcal{O}(K|\mathcal{E}|C)$, where $K$ represents the kernel size, $|\mathcal{E}|$ the number of edges, and $C$ the number of channels. This complexity arises primarily from the polynomial approximation of the graph Laplacian. In contrast, the spatiotemporal self-attention layer demonstrates a more intricate computational profile, with a complexity of $\mathcal{O}(TN^2C + NT^2C)$, where $T$ denotes the number of frames. This increased complexity stems from the dynamic weight generation in multi-head self-attention, encompassing operations such as query-key interactions, softmax computations, and attention-weighted aggregations across both spatial and temporal dimensions. The self-attention approach, while more computationally intensive, offers enhanced flexibility in capturing complex spatiotemporal dependencies, particularly when dealing with lengthy sequences or high-dimensional feature spaces. The choice between these methods thus presents a trade-off between computational efficiency and model expressiveness, contingent upon the specific requirements of the spatiotemporal modeling task at hand.
\begin{figure*}[t]
	\centering
	\includegraphics[width=1\linewidth]{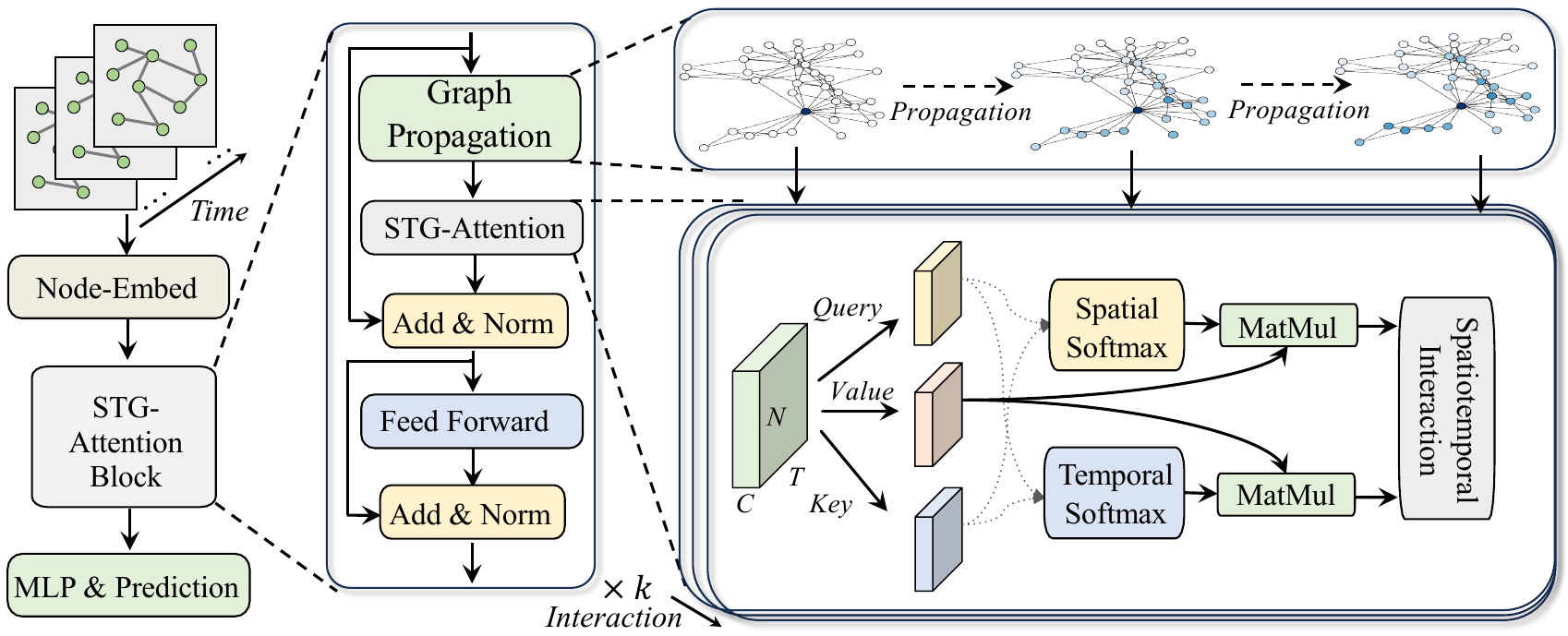}
	\caption{\textbf{The overall architecture of STGformer}. STGformer differs from the traditional Transformer in that we replace the attention mechanism with a spatiotemporal graph attention module, while simplifying the interaction process to require only a single layer rather than stacking multiple layers. Specifically, node features are first propagated through the graph propagation module for \( k \) iterations. Then, the information from each propagation step is fed into the spatiotemporal attention module, where each module shares the same query, key, and values. Finally, the attention outputs are sequentially processed from lower to higher order for interaction.}
	\label{fig:framework}
\end{figure*}
\section{Methodology}
\subsection{Overview}\label{subsec:overview}
The overall architecture of our proposed STGformer is illustrated in \figref{fig:framework}. The key feature of our model is its efficiency, as it achieves joint spatiotemporal graph attention using only a single attention module. Our model is divided into two branches: the graph propagation module and the attention module, with specific details shown in \figref{fig:fig1}. First, the spatiotemporal data undergoes graph propagation and is then fed into the attention module separately. Subsequently, a 1x1 convolution is applied to interact with the outputs of different-order attentions, which are finally aggregated.

\subsection{Data Embedding Layer} To transform the input data into a high-dimensional representation, we adopt a data embedding layer consistent with the STAEformer. Specifically, the raw input \(\mathbf{X}\) is first projected into \(X_{data} \in \mathbb{R}^{T \times N \times d}\) through a fully connected layer, where \(d\) is the embedding dimension.
Recognizing the inherent periodicity of urban traffic flow influenced by human commuting patterns and lifestyles, such as rush hours, we introduce two embeddings to capture weekly and daily cycles, denoted as \(t_{w(t)}, t_{d(t)} \in \mathbb{R}^{d}\), respectively. Here, \(w(t)\) and \(d(t)\) are functions that map time \(t\) to the corresponding week index (1 to 7) and minute index (1 to 288, with a 5-minute interval). The temporal cycle embeddings \(X_w, X_d \in \mathbb{R}^{T \times d}\) are obtained by concatenating the embeddings of all \(T\) time steps. Following \cite{liu2023spatio}, we also incorporate spatiotemporal positional encoding \(X_{ste} \in \mathbb{R}^{N \times T \times d}\) to introduce positional information into the input sequence.
Finally, the output of the data embedding layer is obtained by simply concatenating the aforementioned embedding vectors:
\( X_{emb} = X_{data} \ || \ X_{w} \ || \ X_{d} \ || \ X_{ste}.\)

\subsection{Spatiotemporal Graph Transformer} 
As previously elucidated, GCNs excel in modeling locally high-interaction information, whereas Transformers are adept at capturing global,  limited interaction information. Although Transformer-based methodologies have achieved state-of-the-art performance in traffic forecasting, their quadratic computational complexity with respect to graph size significantly impedes their application, particularly on real-world road networks characterized by high sensor density \cite{chen2001freeway,liu2024largest}. In this study, we propose a more efficient and effective approach to spatiotemporal interaction modeling by synergistically combining graph propagation and Transformer architectures.
We adopt a simplified variant of the GCN formulation presented in Equation \ref{eqn:gcn}, focusing exclusively on graph propagation. For traffic signals, denoted as $X_{emb}$, we omit the feature parameter $W$, resulting in the following formulation:
\begin{equation*}
	\left[ X_0 \ |\  X_1 \ | X_2 \ | \dots \ | \ X_k \right] = \texttt{GraphPropagation}(X_{emb}),
\end{equation*}
where $X_0 = X$ and $X_k = \mathcal{L}_kX$. The \texttt{GraphPropagation} operation is analogous to the simplification introduced by the simplified graph convolution \cite{wu2019simplifying}, which streamlines graph convolutional networks by eliminating nonlinearities and collapsing weight matrices across consecutive layers. However, in contrast to SGC, our approach retains the different orders of $X_k$ to facilitate further interactions.

We then propose a recursive attention module to introduce higher-order spatiotemporal interactions, further enhancing the model's capability. The recursive attention module first takes graph-propagated information as input, then recursively applies gated convolutions: 
\begin{align}
	p_{n+1} = a_n(q_n) \odot g_n(p_n), 
\end{align}
where $k = 0, 1, \ldots, n-1$, with $a_k$ representing the spatiotemporal attention module and $g_k$ used for dimensional matching:
\begin{align}
	g_n= \begin{cases}\text { Identity, } & n=0 \\ \text { Linear }\left(C_{n-1}, C_n\right), & 1 \leq n \leq k-1\end{cases}
\end{align}
As mentioned earlier in \secref{sec:stae}, traditional spatiotemporal attention mechanisms mainly capture spatiotemporal patterns through a separable approach and achieve higher-order interactions by stacking layers. In contrast, as shown in \figref{fig:framework}, we treat space and time as a unified entity, employing a single projection to generate the query, key, and value vectors, and using a simple transposition to compute the attention mechanism. $Q = h w_Q, \ K = h w_K, \ V = h w_V,$
where \( w_Q, w_K, w_V \in \mathbb{R}^{C \times C} \) are trainable parameters. Subsequently, the spatial and temporal self-attention scores are defined as:
\begin{align}\label{eqn:att_slow}
	A^{s} = \text{Softmax}\left(\frac{Q K^{T}}{\sqrt{C}}\right), \quad A^{t} = \text{Softmax}\left(\frac{Q^{T} K}{\sqrt{C}}\right),
\end{align}
where \( A^{s} \in \mathbb{R}^{T \times N \times N} \) captures spatial relations across different nodes, and \( A^{t} \in \mathbb{R}^{N \times T \times T} \) captures temporal relations within individual nodes.
However, despite our integration of spatiotemporal attention, which partially reduces the computational overhead, the inherent quadratic computational complexity still results in considerable computational costs. Therefore, we will next introduce a linear attention mechanism to further alleviate this issue.


\noindent\textbf{Spatiotemporal Linearized Attention.}
Recent work \cite{liu2024largest} have point out that a major part of existing both graph neural network (GCN) and Transformer-based method failed to adopt to the real road graph since the high computation cost in capture global and local traffic pattern. We consider the major issue is the high computation cost with graph size in real world increase dramatically. To address this challenge, the efficient attention mechanism \cite{katharopoulos2020Transformers,wang2020linformer,wu2024simplifying} is adopt in this paper to address the significant resource demands of traditional dot-product attention, which has quadratic memory and computational complexities. This makes dot-product attention impractical for real world traffic graph due to its high computational and memory costs.
Efficient attention maintains mathematical equivalence to dot-product attention but achieves substantial improvements in speed and memory efficiency, which is done by interpreting the keys differently: instead of viewing them as \( N \) node feature vectors in \( \mathbb{R}^{d} \), they are treated as \( d \) single-channel feature maps. Each of these maps acts as a weighting over all positions, aggregating value features through weighted summation to form a global context vector. The core equation for efficient attention is:
\begin{align*}
\bm{E}(Q, K, V) = w_q(X)\left(w_k(X)^\mathsf{T}w_v(X)\right),
\end{align*}
where \(w_q\) and \(w_k\) are normalization functions. For scaling, both \(w_q\) and \(w_k\) divide the matrices by \(\sqrt{n}\). 
We can easily prove the equivalence between dot-product and efficient attention with scaling normalization demonstrates that:
\begin{align}\label{eqn:fast}
\bm{E}(Q, K, V) = \frac{Q}{\sqrt{n}}\left(\frac{K^\mathsf{T}}{\sqrt{n}}V\right) = \frac{1}{n}(QK^\mathsf{T})V,
\end{align}
Therefore, we substitute \eqnref{eqn:att_slow} into \eqnref{eqn:fast} and get
\begin{align}\label{eqn:att_fast}
	A^{s} = \frac{1}{n}(QK^\mathsf{T})V, \quad A^{t} = \frac{1}{n}(Q^\mathsf{T}K)V,
\end{align}

\subsection{Computational Cost Analysis.}
From \eqnref{eqn:att_slow}, it is evident that the computational cost of the spatiotemporal softmax attention mechanism scales with $\mathcal{O}\left(TN^2 + NT^2\right)$. The memory requirements similarly increase, as the complete attention matrix must be stored to compute the gradients for the queries, keys, and values. In contrast, the linear transformer we adopt in \eqnref{eqn:att_fast} has both time and memory complexities of $\mathcal{O}(N + T)$. 

We will divide the computation of our into 3 parts, and calculate the FLOPs for each part.

\begin{itemize} 
    \item[$\bullet$] \textbf{Graph Propagation.} Graph propagation, employing Chebyshev polynomials, owns  a computational complexity of \(\mathcal{O}(K|\mathcal{E}|C)\), where \(K\) denotes the order, \(|\mathcal{E}|\) represents the number of edges, and \(C\) signifies the number of channels. This complexity is predominantly attributed to the polynomial approximation of the graph Laplacian.
    \item[$\bullet$] \textbf{Spatiotemporal Linear Attention.} As previously mentioned, the time complexity of Spatiotemporal Linear Attention is \(\mathcal{O}(N + T)\), where \(N\) is the number of spatial nodes and \(T\) is the temporal length. Since the process needs to be performed \(K\) times, the overall computational complexity becomes \(\mathcal{O}(K(N + T))\).
    \item[$\bullet$] \textbf{Recursive Interaction.}   We consider  the FLOPs of the element-wise multiplication with 1$\times$1 convolution. Therefore, the computational cost is $KNTC^2$.
\end{itemize}
Therefore, the total FLOPs with spatiotemporal attention are:
\begin{align*}
	\flops(\text{STGformer}) &= KC(|\mathcal{E}| + N + T + NTC).
\end{align*}
Because STAEFormer performs self-attention operations on spatial and temporal stacked with $L$ layers respectively, we can easily calculate its FLOPs as:
\begin{align*}
	\flops(\text{STAEFormer}) &= L(TN^2·C + NT^2·C).
\end{align*}
For instance, assuming input lengths of 12, California graph with 8600 nodes, a hidden dimension of 32,  an interaction order \(K\) of 3, $|\mathcal{E}|=201,363$ and $L = 3$, the ratio of FLOPs between STGformer and STAEFormer can be calculated as follows:
\[
\frac{\text{FLOPs(STGformer)}}{\text{FLOPs(STAEFormer)}} \approx 0.00131,
\]
which STGformer significantly reduces  99.869\% computational burden compared to STAEFormer.

\begin{table*}[t]
	\caption{\textbf{Performance comparisons are presented, with the best-performing baseline results highlighted in bold.}  "Param" denotes the number of learnable parameters, where K represents thousands ($10^3$) and M represents millions ($10^6$).}
	\label{tab:performance}
	\centering
	\small
	\renewcommand{\arraystretch}{1.3}
	\begin{sc}
		\resizebox{\textwidth}{!}{
			\begin{tabular}{lccccc|ccc|ccc|ccc}
				\toprule
				\multirow{2}{*}{Data} & \multirow{2}{*}{Method} & \multirow{2}{*}{Param} & \multicolumn{3}{c}{Horizon 3} & \multicolumn{3}{c}{Horizon 6} & \multicolumn{3}{c}{Horizon 12} & \multicolumn{3}{c}{Average} \\ \cline{4-15} 
				&  &  & MAE & RMSE & MAPE & MAE & RMSE & MAPE & MAE & RMSE & MAPE & MAE & RMSE & MAPE \\ 
				\hline \hline
				\multirowcell{13}{San Diego\\($N = 716$)} & HA & -- & 33.61 & 50.97 & 20.77\% & 57.80 & 84.92 & 37.73\% & 101.74 & 140.14 & 76.84\% & 60.79 & 87.40 & 41.88\% \\
				& LSTM & 98K & 19.03 & 30.53 & 11.81\% & 25.84 & 40.87 & 16.44\% & 37.63 & 59.07 & 25.45\% & 26.44 & 41.73 & 17.20\% \\
				& DCRNN & 373K & 17.14 & 27.47 & 11.12\% & 20.99 & 33.29 & 13.95\% & 26.99 & 42.86 & 18.67\% & 21.03 & 33.37 & 14.13\% \\
				& AGCRN & 761K & 15.71 & 27.85 & 11.48\% & 18.06 & 31.51 & 13.06\% & 21.86 & 39.44 & 16.52\% & 18.09 & 32.01 & 13.28\% \\
				& STGCN & 508K & 17.45 & 29.99 & 12.42\% & 19.55 & 33.69 & 13.68\% & 23.21 & 41.23 & 16.32\% & 19.67 & 34.14 & 13.86\% \\
				& GWNET & 311K & 15.24 & 25.13 & 9.86\% & 17.74 & 29.51 & 11.70\% & 21.56 & 36.82 & 15.13\% & 17.74 & 29.62 & 11.88\% \\
				& ASTGCN & 2.2M & 19.56 & 31.33 & 12.18\% & 24.13 & 37.95 & 15.38\% & 30.96 & 49.17 & 21.98\% & 23.70 & 37.63 & 15.65\% \\
				& STGODE & 729K & 16.75 & 28.04 & 11.00\% & 19.71 & 33.56 & 13.16\% & 23.67 & 42.12 & 16.58\% & 19.55 & 33.57 & 13.22\% \\
				& DSTAGNN & 3.9M & 18.13 & 28.96 & 11.38\% & 21.71 & 34.44 & 13.93\% & 27.51 & 43.95 & 19.34\% & 21.82 & 34.68 & 14.40\% \\
				& DGCRN & 243K & 15.34 & 25.35 & 10.01\% & 18.05 & 30.06 & 11.90\% & 22.06 & 37.51 & 15.27\% & 18.02 & 30.09 & 12.07\% \\
				& D$^2$STGNN & 406K & 14.92 & 24.95 & 9.56\% & 17.52 &  29.24 & 11.36\% & 22.62 & 37.14 & 14.86\% & 17.85 & 29.51 &  11.54\% \\
				&STID &258K& 15.08& 25.20& 9.88\%& 17.79& 30.15& 11.97\%& 21.68& 38.59& 15.15\%& 17.82& 30.98& 11.96\% \\
				& STAEformer & 1.7M & 15.37&  25.66&  10.15\%& 18.03&  30.46&  12.11\%&22.21&  37.79&  15.49\%&18.01 & 30.38&  12.03\%\\
				\rowcolor{Gray} &  STGformer & 256K & \bf 14.97 & \bf 24.96 & \bf 9.41\% & \bf 17.44 & \bf 29.26 & \bf 11.12\% & \bf 20.94& \bf 35.93 & \bf 14.08\% & \bf 17.36 & \bf 29.52 & \bf 11.22\% \\
				\hline \hline
				\multirowcell{11}{Bay Area\\($N = 2,352$)} & HA & -- & 32.57 & 48.42 & 22.78\% & 53.79 & 77.08 & 43.01\% & 92.64 & 126.22 & 92.85\% & 56.44 & 79.82 & 48.87\% \\
				& LSTM & 98K & 20.38 & 33.34 & 15.47\% & 27.56 & 43.57 & 23.52\% & 39.03 & 60.59 & 37.48\% & 27.96 & 44.21 & 24.48\% \\
				& DCRNN & 373K & 18.71 & 30.36 & 14.72\% & 23.06 & 36.16 & 20.45\% & 29.85 & 46.06 & 29.93\% & 23.13 & 36.35 & 20.84\% \\
				& AGCRN & 777K & 18.31 & 30.24 & 14.27\% & 21.27 & 34.72 & 16.89\% &  24.85 & 40.18 & 20.80\% & 21.01 & 34.25 & 16.90\% \\
				& STGCN & 1.3M & 21.05 & 34.51 & 16.42\% & 23.63 & 38.92 & 18.35\% & 26.87 & 44.45 & 21.92\% & 23.42 & 38.57 & 18.46\% \\
				& GWNET & 344K & 17.85 & 29.12 & 13.92\% & 21.11 & 33.69 & 17.79\% & 25.58 & 40.19 & 23.48\% & 20.91 & 33.41 & 17.66\% \\
				& ASTGCN & 22.3M & 21.46 & 33.86 & 17.24\% & 26.96 & 41.38 & 24.22\% & 34.29 & 52.44 & 32.53\% & 26.47 & 40.99 & 23.65\% \\
				& STTN & 218K & 18.25 & 29.64 & 14.05\% & 21.06 & 33.87 & 17.03\% & 25.29 & 40.58 & 21.20\% & 20.97 & 33.78 & 16.84\% \\
				& STGODE & 788K & 18.84 & 30.51 & 15.43\% & 22.04 & 35.61 & 18.42\% & 26.22 & 42.90 & 22.83\% & 21.79 & 35.37 & 18.26\% \\
				& DSTAGNN & 26.9M & 19.73 & 31.39 & 15.42\% & 24.21 & 37.70 & 20.99\% & 30.12 & 46.40 & 28.16\% & 23.82 & 37.29 & 20.16\% \\
				& DGCRN & 374K & 18.02 & 29.49 & 14.13\% & 21.08 & 34.03 & 16.94\% & 25.25 & 40.63 & 21.15\% & 20.91 & 33.83 & 16.88\% \\
				& STID &711K& 17.25& 29.18& 13.42\%& 20.31& 34.20& 16.13\%& 24.29& 41.29& 20.16\%& 20.14& 34.39& 16.07\% \\
				& D$^2$STGNN & 446K & 17.54 & 28.94 & 12.12\% & 20.92 &  33.92 & 14.89\% & 25.48 & 40.99 &  19.83\% & 20.71 & 33.65 &  15.04\% \\
				& STAEformer & 3.3M & 17.55 & 29.25 & 13.00\% & 20.55 & 33.87 & 15.45\% & 24.75 & 41.00 & 19.75\% & 20.39 & 34.21 & 15.55\% \\
				\rowcolor{Gray} & STGformer &491K  &  \bf  17.13 & \bf 28.63 & \bf 12.72\% &\bf 20.11 & \bf 33.20&\bf 15.12\%  &\bf 24.22&\bf 40.16 &\bf 19.35\% &\bf 19.98 &\bf 33.50 & \bf 15.22\% \\
				\hline \hline
				\multirowcell{11}{Los Angeles\\($N = 3,834$)} & HA & -- & 33.66 & 50.91 & 19.16\% & 56.88 & 83.54 & 34.85\% & 98.45 & 137.52 & 71.14\% & 59.58 & 86.19 & 38.76\% \\
				& LSTM & 98K & 20.02 & 32.41 & 11.36\% & 27.73 & 44.05 & 16.49\% & 39.55 & 61.65 & 25.68\% & 28.05 & 44.38 & 17.23\% \\
				& DCRNN & 373K & 18.41 & 29.23 & 10.94\% & 23.16 & 36.15 & 14.14\% & 30.26 & 46.85 & 19.68\% & 23.17 & 36.19 & 14.40\% \\
				& AGCRN & 792K & 17.27 & 29.70 & 10.78\% &  20.38 & 34.82 & 12.70\% &  24.59 & 42.59 & 16.03\% & 20.25 & 34.84 & 12.87\% \\
				& STGCN & 2.1M & 19.86 & 34.10 & 12.40\% & 22.75 & 38.91 & 14.11\% & 26.70 & 45.78 & 17.00\% & 22.64 & 38.81 & 14.17\% \\
				& GWNET & 374K & 17.28 & 27.68 & 10.18\% & 21.31 & 33.70 & 13.02\% & 26.99 & 42.51 & 17.64\% & 21.20 & 33.58 & 13.18\% \\
				& ASTGCN & 59.1M & 21.89 & 34.17 & 13.29\% & 29.54 & 45.01 & 19.36\% & 39.02 & 58.81 & 29.23\% & 28.99 & 44.33 & 19.62\% \\
				& STGODE & 841K & 18.10 & 30.02 & 11.18\% & 21.71 & 36.46 & 13.64\% & 26.45 & 45.09 & 17.60\% & 21.49 & 36.14 & 13.72\% \\
				& DSTAGNN & 66.3M & 19.49 & 31.08 & 11.50\% & 24.27 & 38.43 & 15.24\% & 30.92 & 48.52 & 20.45\% & 24.13 & 38.15 & 15.07\% \\
				&STID &901K& 16.43& 27.40& 9.89\%& 19.77& 33.43& 12.26\%& 24.23& 42.02& 15.88\%& 19.66& 33.99& 12.31\%\\
				& STAEformer & 4.7M & 16.72 & 27.50 & 9.77\% & 20.10 & 33.05 & 11.89\% & 24.69 & 41.42 & 15.47\% & 19.97 & 33.53 & 12.01\% \\
				\rowcolor{Gray} & STGformer &  705K & \bf 16.39 & \bf 26.95 & \bf 9.58\% & \bf 19.70 & \bf 32.39 & \bf 11.66\% & \bf 24.19 & \bf 40.59 & \bf 15.16\% & \bf 19.58 & \bf 32.88 & \bf 11.78\% \\
				\bottomrule \bottomrule
			\end{tabular}
		}
	\end{sc}
\end{table*}
\section{Experiment}
\noindent\textbf{Datasets.} We experimented with LargeST \cite{liu2024largest}, which aggregated traffic readings from 5-minute intervals into 15-minute windows, aiming to predict future 12-step outcomes based on historical 12-step data \cite{jiang2021dl}.  LargeST comprises three California sub-datasets constructed from three representative regions within the state. The first is Los Angeles, encompassing 3,834 sensors installed across five counties in the Los Angeles region: Los Angeles, Orange, Riverside, San Bernardino, and Ventura. The second sub-dataset, the Bay Area, includes 2,352 sensors located in 11 counties: Alameda, Contra Costa, Marin, Napa, San Benito, San Francisco, San Mateo, Santa Clara, Santa Cruz, Solano, and Sonoma. The smallest sub-dataset, San Diego, contains 716 sensors. To further verify the performance of our method, we also conducted experiments on the widely-used PEMS-series benchmarks i.e., PEMS03, PEMS04, PEMS07, PEMS08.  \cite{guo2019attention}. PEMS-series, representing the four major districts in California, are aggregated into 5-minute intervals, resulting in 12 data points per hour and 288 data points per day.
 
\noindent\textbf{Implementation Details.} Our experiments ran on a GPU server with eight GeForce GTX 3090 graphics cards, employing the PyTorch 2.0.3 framework. Raw data have been standardized using z-score normalization \cite{cheadle2003analysis}. If validation error stabilized within 15-20 epochs or ceased after 200 epochs, training halted prematurely, preserving the best model based on validation data \cite{luo2023dynamic}. We maintained fidelity to the original paper's model parameters and settings, while also conducting multiple parameter tuning iterations to enhance experimental outcomes. Data were partitioned chronologically into training, validation, and test sets at a 6:2:2 ratio across all sub-datasets. In our experiments, we assess model performance using the Mask-Based Root Mean Square Error (RMSE), Mean Absolute Error (MAE), and Mean Absolute Percentage Error (MAPE) as metrics, wherein zero values (indicating noisy data) are disregarded.

\begin{table*}[t]
	\centering
	\small
	\renewcommand{\arraystretch}{1.4}
	\caption{Performance comparisons on cross year scenario, with the best-performing baseline results highlighted in bold.}
	\label{tab:ood}
	\begin{sc}
		\resizebox{\textwidth}{!}{
			\begin{tabular}{lcccc|ccc|ccc|ccc}
				\toprule
				\multirow{2}{*}{Data} & \multirow{2}{*}{Method} & \multicolumn{3}{c}{Horizon 3} & \multicolumn{3}{c}{Horizon 6} & \multicolumn{3}{c}{Horizon 12} & \multicolumn{3}{c}{Average} \\ \cline{3-14} 
				&  &  MAE & RMSE & MAPE & MAE & RMSE & MAPE & MAE & RMSE & MAPE & MAE & RMSE & MAPE \\ 
				\hline \hline
				\multirowcell{13}{San Diego\\ 2019 \\ $\downarrow$ \\ 2020}
				& AGCRN & 21.36 & 34.17 & 26.36\% & 27.37 & 42.02 & 32.45\% & 35.14 & 52.60 & 42.91\% & 27.18 & 41.90 & 32.95\% \\
				& ASTGCN & 20.44 & 32.75 & 17.91\% & 31.33 & 51.44 & 28.05\% & 40.38 & 65.34 & 41.34\% & 30.08 & 48.98 & 27.10\% \\
				& D2STGNN & 19.52 & 30.84 & 20.86\% & 25.32 & 38.85 & 27.68\% & 33.19 & 48.55 & 34.87\% & 25.33 & 38.40 & 27.81\% \\
				& DGCRN & 18.17 & 27.89 & 18.20\% & 24.43 & 37.42 & 27.77\% & 39.34 & 58.15 &\bf 35.98\% & 25.63 & 38.88 & 27.50\% \\
				& DSTAGNN & 20.89 & 33.13 & 18.06\% & 30.26 & 46.25 & 28.82\% & 41.15 & 60.90 & 41.99\% & 29.65 & 45.17 & 28.17\% \\
				& GWNET & 18.16 & 29.01 & 17.38\% & 24.52 & 38.55 & 27.75\% & 32.57 & 47.92 & 37.63\% & 24.58 & 37.61 & 27.89\% \\
				& STGCN & 28.48 & 41.99 & 36.03\% & 33.44 & 48.43 & 39.78\% & 38.55 & 56.53 & 40.97\% & 32.90 & 48.11 & 38.78\% \\
				& STGODE & 20.29 & 31.67 & 21.60\% & 27.14 & 41.31 & 29.82\% & 33.52 & 49.51 & 38.64\% & 26.29 & 39.69 & 29.28\% \\
				& STID & 18.38 & 29.12 & 17.90\% & 25.00 & 38.13 & 27.75\% & 32.65 & 48.33 & 38.08\% & 24.52 & 38.29 &\bf 26.88\% \\
				& STAEformer & 21.33 & 31.55 & 29.07\% & 26.66 & 39.20 & 33.41\% & 32.98 & 47.53 & 38.65\% & 26.22 & 38.29 & 32.74\% \\
				\rowcolor{Gray} & STGformer & \bf 17.91 & \bf 27.09 &  \bf 17.20\% & \bf 24.19 & \bf 37.09 &  \bf 27.64\% & \bf 31.94 & \bf 47.26 &  37.53\% & \bf 23.92 & \bf 37.30 &  27.54\% \\
				\bottomrule 
				\multirowcell{13}{Bay Area \\ 2019 \\ $\downarrow$ \\ 2020}
				& AGCRN & 23.46 & 35.28 & 19.93\% & 31.44 & 46.63 & 25.30\% & 39.87 & 57.83 & 32.48\% & 30.71 & 45.33 & 25.12\% \\
				& ASTGCN & 22.54 & 33.86 & 18.48\% & 33.43 & 49.05 & 26.90\% & 42.38 & 61.57 & 34.91\% & 32.18 & 47.41 & 26.23\% \\
				& D2STGNN & 21.62 & 31.95 & 17.43\% & 27.42 & 41.46 & 23.53\% & 35.19 & 51.78 & \bf 28.44\% & 27.43 & 40.83 & 22.84\% \\
				& DGCRN & 20.27 & 30.00 & 16.77\% & 26.53 & 40.03 & 23.62\% & 41.34 & 55.38 & 29.98\% & 27.73 & 40.81 & 23.55\% \\
				& DSTAGNN & 22.99 & 34.24 & 18.63\% & 32.36 & 48.86 & 24.67\% & 43.15 & 57.13 & 35.49\% & 31.75 & 46.60 & 26.22\% \\
				& GWNET & 20.26 & 30.12 & 16.95\% & 26.62 & 41.16 & 23.60\% & 34.57 & 52.15 & 31.63\% & 26.58 & 40.04 & 23.89\% \\
				& STGCN & 30.36 & 43.13 & 26.55\% & 35.47 & 50.06 & 29.31\% & 40.40 & 56.84 & 32.35\% & 34.80 & 49.18 & 29.07\% \\
				& STGODE & 22.39 & 32.78 & 18.60\% & 29.24 & 43.92 & 25.67\% & 35.52 & 53.74 & 32.64\% & 28.29 & 42.12 & 25.30\% \\
				& STID & 19.40 & 29.69 & 14.86\% & 27.09 & 40.15 & \bf 22.17\% & 36.08 & 52.10 & 30.51\% & 26.55 & 39.30 &\bf 21.68\% \\
				& STAEformer & 23.43 & 32.66 & 23.07\% & 28.76 & 41.81 & 29.26\% & 34.98 & 51.76 & 32.65\% & 28.22 & 40.72 & 27.74\% \\
				\rowcolor{Gray} & STGformer & \bf 19.01 & \bf 28.20 & \bf 14.20\% & \bf 26.29 & \bf 39.70 &  23.49\% & \bf 33.94 & \bf 51.49 &  31.53\% & \bf 25.92 & \bf 39.73 & 22.54\% \\
				\hline \hline
				\multirowcell{13}{Los Angeles \\ 2019 \\ $\downarrow$ \\ 2020}
				& AGCRN & 22.85 & 36.59 & 22.69\% & 31.66 & 48.33 & 30.95\% & 41.21 & 61.71 & 40.04\% & 31.13 & 47.80 & 30.38\% \\
				& ASTGCN & 21.93 & 35.17 & 19.24\% & 33.65 & 50.75 & 29.55\% & 43.72 & 65.45 & 42.47\% & 32.60 & 49.88 & 29.49\% \\
				& D2STGNN & 21.01 & 33.26 & 21.19\% & 27.64 & 43.16 & 28.18\% & 36.53 & 55.66 & 35.97\% & 27.85 & 43.30 & 28.31\% \\
				& DGCRN & 19.66 & 31.31 & 18.53\% & 26.75 & 41.73 & 28.27\% & 42.68 & 59.26 & 37.08\% & 28.15 & 43.28 & 28.00\% \\
				& DSTAGNN & 22.38 & 35.55 & 18.39\% & 32.58 & 49.56 & 29.32\% & 44.49 & 61.01 & 43.09\% & 32.17 & 47.97 & 28.67\% \\
				& GWNET & 19.20 & 29.69 & 15.20\% & 26.94 & 40.31 & 22.14\% & 38.34 & 56.90 & 32.52\% & 27.20 & 40.85 & 22.51\% \\
				& STGCN & 30.42 & 46.30 & 34.14\% & 36.51 & 54.42 & 38.00\% & 42.62 & 63.34 & 42.18\% & 35.85 & 53.74 & 37.58\% \\
				& STGODE & 21.78 & 34.09 & 21.93\% & 29.46 & 45.62 & 30.32\% & 36.86 & 57.62 & 39.74\% & 28.71 & 45.05 & 29.78\% \\
				& STID & 20.32 & 31.52 & 17.23\% & 29.07 & 43.62 & 27.97\% & 39.67 & 58.29 & 40.73\% & 28.71 & 43.10 & 27.69\% \\
				& STAEformer & 22.82 & 33.97 & 29.40\% & 28.98 & 43.51 & 33.91\% & 36.32 & 55.64 & 39.75\% & 28.64 & 43.65 & 33.24\% \\
				\rowcolor{Gray} & STGformer & \bf 19.40 & \bf 29.52 & \bf 17.53\% & \bf 26.51 & \bf 41.40 & \bf 28.14\% & \bf 35.28 & \bf 55.37 & \bf 38.63\% & \bf 26.34 & \bf 41.76 & \bf 27.04\% \\
				\bottomrule \bottomrule
			\end{tabular}
		}
	\end{sc}
\end{table*}

\begin{table*}[t]
	\centering
	\caption{\textbf{Performance on PEMS03, PEMS04, PEMS07, and PEMS08 benchmarks w/o using incident information.} }
	\label{tab:perf_pems}
	\renewcommand\arraystretch{1.3}
	\begin{sc}
		\resizebox{\linewidth}{!}{%
			\begin{tabular}{ccc|ccccccccc>{\columncolor{Gray}}c}
				\toprule
				\multicolumn{2}{c}{Dataset}                                                                          & Metric & HA      & STGCN   & DCRNN   & GWNet   & AGCRN   & GTS     & STID    & PDFormer & STAEformer &  STGformer \\
				\hline
				\multirow{3}{*}{\rotatebox{90}{\small PEMS03}}  & \multirow{3}{*}{Average}                                                  & MAE    & 32.62   & 15.83   & 15.54   & 14.59   & 15.24   & 15.41   & 15.33   & 14.94    & 15.35      & \bf 14.47   \\
				&                                                                           & RMSE   & 49.89   & 27.51   & 27.18   & 25.24   & 26.65   & 26.15   & 27.40   & 25.39    & 27.55      & \bf 25.08   \\
				&                                                                           & MAPE   & 30.60\% & 16.13\% & 15.62\% & 15.52\% & 15.89\% & 15.39\% & 16.40\% & 15.82\%  & 15.18\%    & \bf 14.41\%   \\
				\midrule
				\multirow{3}{*}{\rotatebox{90}{\small PEMS04}}  & \multirow{3}{*}{Average}                                                  & MAE    & 42.35   & 19.57   & 19.63   & 18.53   & 19.38   & 20.96   & 18.38   & 18.36    & 18.22      & \bf 17.89   \\
				&                                                                           & RMSE   & 61.66   & 31.38   & 31.26   & 29.92   & 31.25   & 32.95   & 29.95   & 30.03    & 30.18      & \bf 29.79   \\
				&                                                                           & MAPE   & 29.92\% & 13.44\% & 13.59\% & 12.89\% & 13.40\% & 14.66\% & 12.04\% & 12.00\%  & 11.98\%    &\bf  11.83\% \\
				\midrule
				\multirow{3}{*}{\rotatebox{90}{\small PEMS07}}  & \multirow{3}{*}{Average}                                                  & MAE    & 49.03   & 21.74   & 21.16   & 20.47   & 20.57   & 22.15   & 19.61   & 19.97    & 19.14      & \bf 19.04   \\
				&                                                                           & RMSE   & 71.18   & 35.27   & 34.14   & 33.47   & 34.40   & 35.10   & 32.79   & 32.95    & 32.60      & \bf 32.50   \\
				&                                                                           & MAPE   & 22.75\% & 9.24\%  & 9.02\%  & 8.61\%  & 8.74\%  & 9.38\%  & 8.30\%  & 8.55\%   & 8.01\%     & \bf 7.89\%  \\
				\midrule
				\multirow{3}{*}{\rotatebox{90}{\small PEMS08}}  & \multirow{3}{*}{Average}                                                  & MAE    & 36.66   & 16.08   & 15.22   & 14.40   & 15.32   & 16.49   & 14.21   & 13.58    & 13.46      & \bf 13.41   \\
				&                                                                           & RMSE   & 50.45   & 25.39   & 24.17   & 23.39   & 24.41   & 26.08   & 23.28   & 23.41    & 23.25      & \bf 23.21   \\
				&                                                                           & MAPE   & 21.63\% & 10.60\% & 10.21\% & 9.21\%  & 10.03\% & 10.54\% & 9.27\%  & 9.05\%   & 8.88\%     & \bf 8.77\% \\
				\bottomrule
			\end{tabular}%
		}
	\end{sc}
\end{table*}

\begin{figure*}[t]
	\centering
	\includegraphics[width=0.8\linewidth]{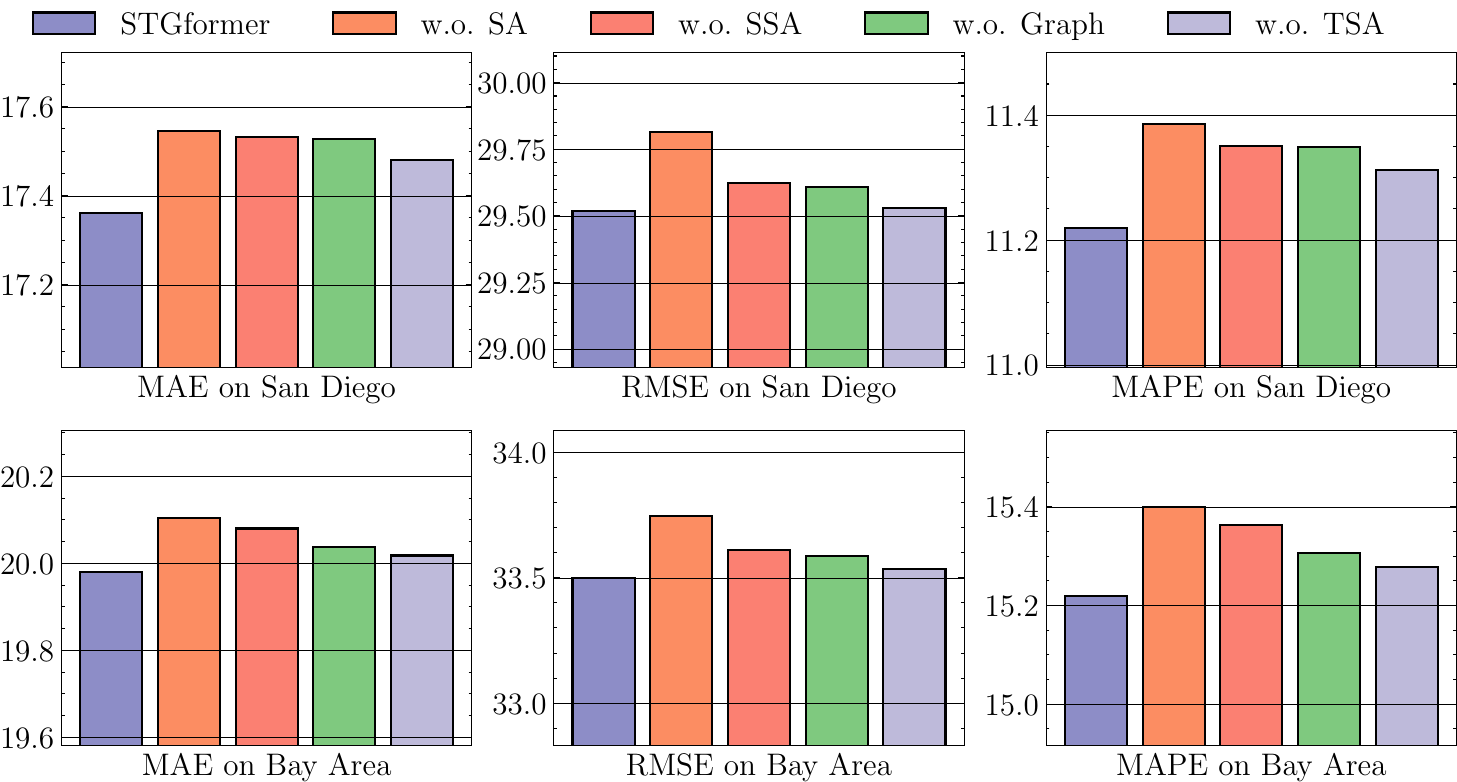}
	\caption{\textbf{Ablation study on San Diego and Bay Area.}}
	\label{fig:abla2}
\end{figure*}

\noindent\textbf{Baselines.} In this study, we conduct a comprehensive evaluation of traffic forecasting methodologies, encompassing a diverse array of baselines with publicly available implementations. These baselines span traditional approaches, contemporary deep learning techniques, and state-of-the-art models, providing a thorough representation of the field's progression. 

\begin{itemize}
	\item \textbf{HA (Historical Average):} Conceptualizes traffic flows as periodic processes, utilizing weighted averages from antecedent periods for future predictions.
	
	\item \textbf{LSTM} \cite{hochreiter1997long}: Long Short-Term Memory, a type of recurrent neural network architecture that is particularly effective at learning and remembering long-term dependencies in sequential data. 
	
	\item \textbf{DCRNN} \cite{li2018diffusion}: Diffusion Convolutional Recurrent Neural Network, which models traffic flow as a diffusion process, innovatively replacing the fully connected layer in Gated Recurrent Units (GRU) \cite{cho2014learning} with a diffusion convolutional layer.
	
	\item \textbf{Graph WaveNet} \cite{wu2019graph}: Strategically stacks Gated Temporal Convolutional Networks (TCN) and Graph Convolutional Networks (GCN) to concurrently capture spatial and temporal dependencies.
	
	\item \textbf{ASTGCN} \cite{guo2019attention}: Attention-based Spatial-Temporal Graph Convolutional Network, which synergistically combines spatial-temporal attention mechanisms to simultaneously capture dynamic spatial-temporal characteristics of traffic data.

	\item \textbf{MTGNN} \cite{wu2020connecting}: Multi-Task Graph Neural Network, which extends the Graph WaveNet through the integration of mix-hop propagation layers in the spatial module, dilated inception layers in the temporal module, and a more sophisticated graph learning layer.
	
	\item \textbf{DGCRN} \cite{li2021dynamic}: Dynamic Graph Convolutional Recurrent Network, which uses hyper-networks to model dynamic spatial relationships and employs an efficient training strategy for improved traffic prediction performance.
	\item \textbf{GTS} \cite{shang2021discrete}: Graph Structure Learning for Time Series, a method that simultaneously learns the graph structure and performs forecasting for multiple time series. 
	\item \textbf{STGCN} \cite{yu2018spatio}: Spatial-Temporal Graph Convolutional Networks, which utilizes graph convolutions to model spatial dependencies and 1D convolutions for temporal modeling in traffic forecasting.
	\item \textbf{STTN} \cite{xu2020spatial}: Spatial-Temporal Transformer Network, which applies the Transformer architecture to capture both spatial and temporal dependencies in traffic data.
	\item \textbf{STGODE} \cite{fang2021spatial}: Spatial-Temporal Graph Ordinary Differential Equation, which uses neural ordinary differential equations to model continuous changes in traffic signals over time and space.
	\item \textbf{DSTAGNN} \cite{lan2022dstagnn}: Dynamic Spatial-Temporal Attention Graph Neural Network, which incorporates dynamic graph learning and attention mechanisms to capture evolving spatial-temporal dependencies in traffic networks.
	\item \textbf{D2STGNN} \cite{shao2022decoupled}: Decoupled Dynamic Spatial-Temporal Graph Neural Network, which separates the modeling of spatial and temporal dependencies while addressing dynamic correlations among sensors in traffic networks.
	\item \textbf{STID} \cite{shao2022spatial}: Spatial-Temporal Identity, a model that combines spatial and temporal embeddings to capture the unique characteristics of each node and time step in traffic forecasting.
	\item \textbf{PDFormer} \cite{jiang2023pdformer}: Propagation Delay-aware Dynamic Long-Range Transformer for traffic flow prediction. It innovatively captures dynamic spatial dependencies, models both short- and long-range spatial information, and explicitly accounts for the time delay in traffic condition propagation between locations.
	\item \textbf{STAEformer} \cite{liu2023spatio}: Spatio-Temporal Adaptive Embedding transformer, enhancing vanilla transformers with adaptive embeddings to capture complex spatio-temporal traffic patterns effectively.
\end{itemize}

\begin{figure*}[t]
	\centering
	\subfigure[Node 24 in San Diego]{\includegraphics[width=0.48\linewidth]{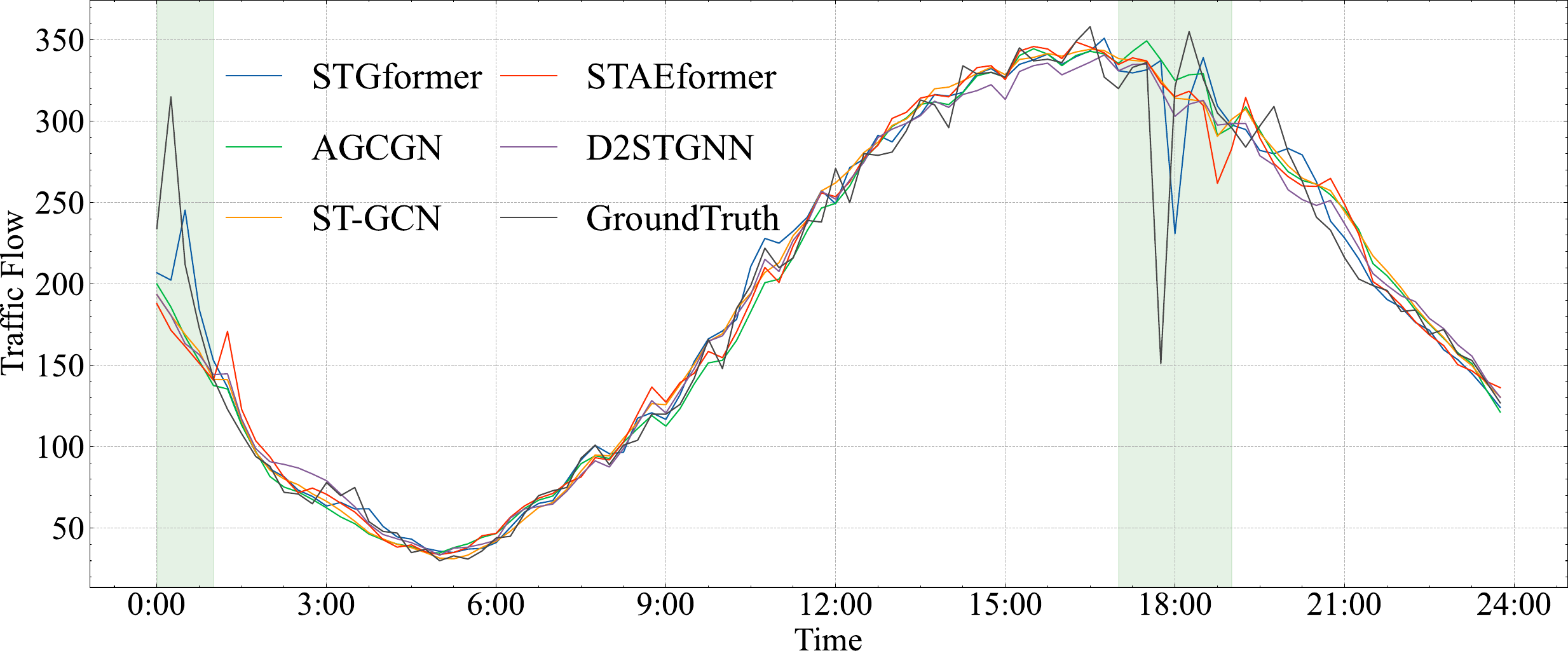}}
	\subfigure[Node 64 in San Diego]{\includegraphics[width=0.48\linewidth]{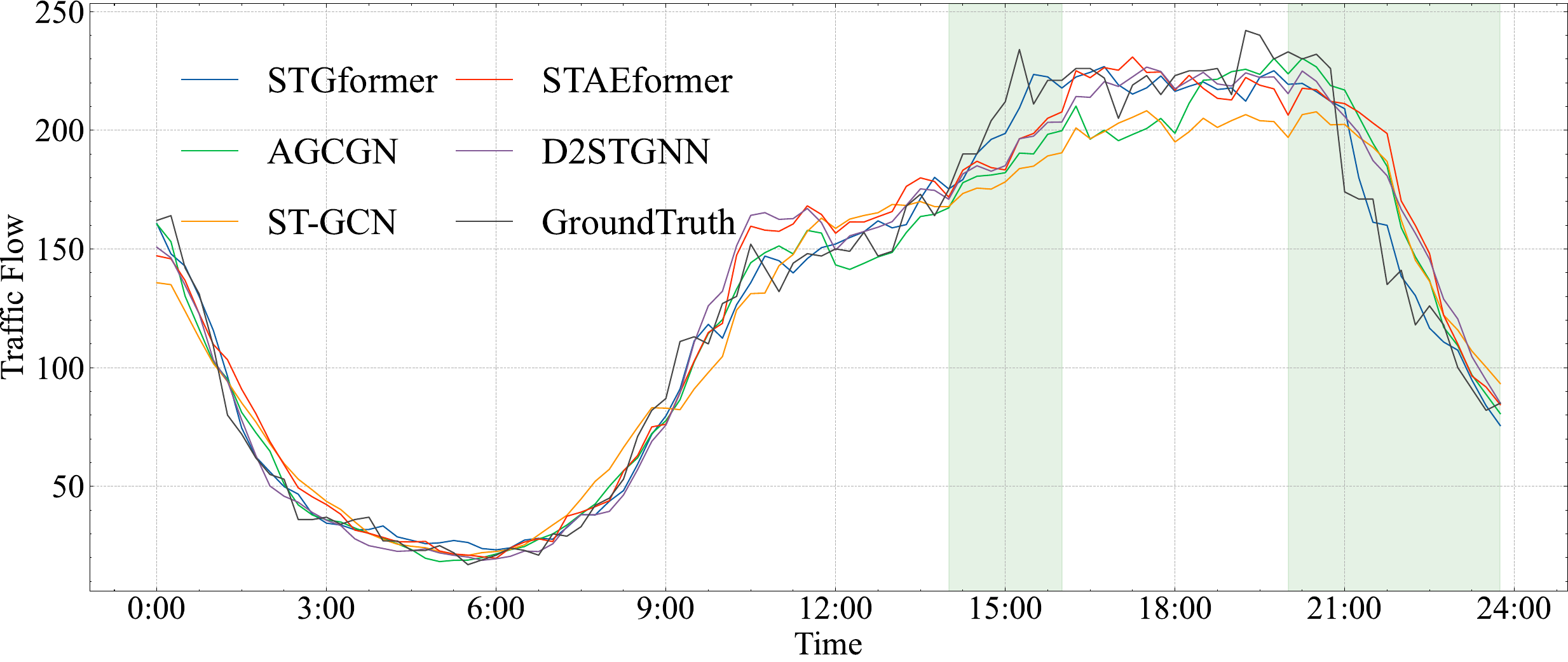}}
	
	\subfigure[Node 93 in  Los Angeles]{\includegraphics[width=0.48\linewidth]{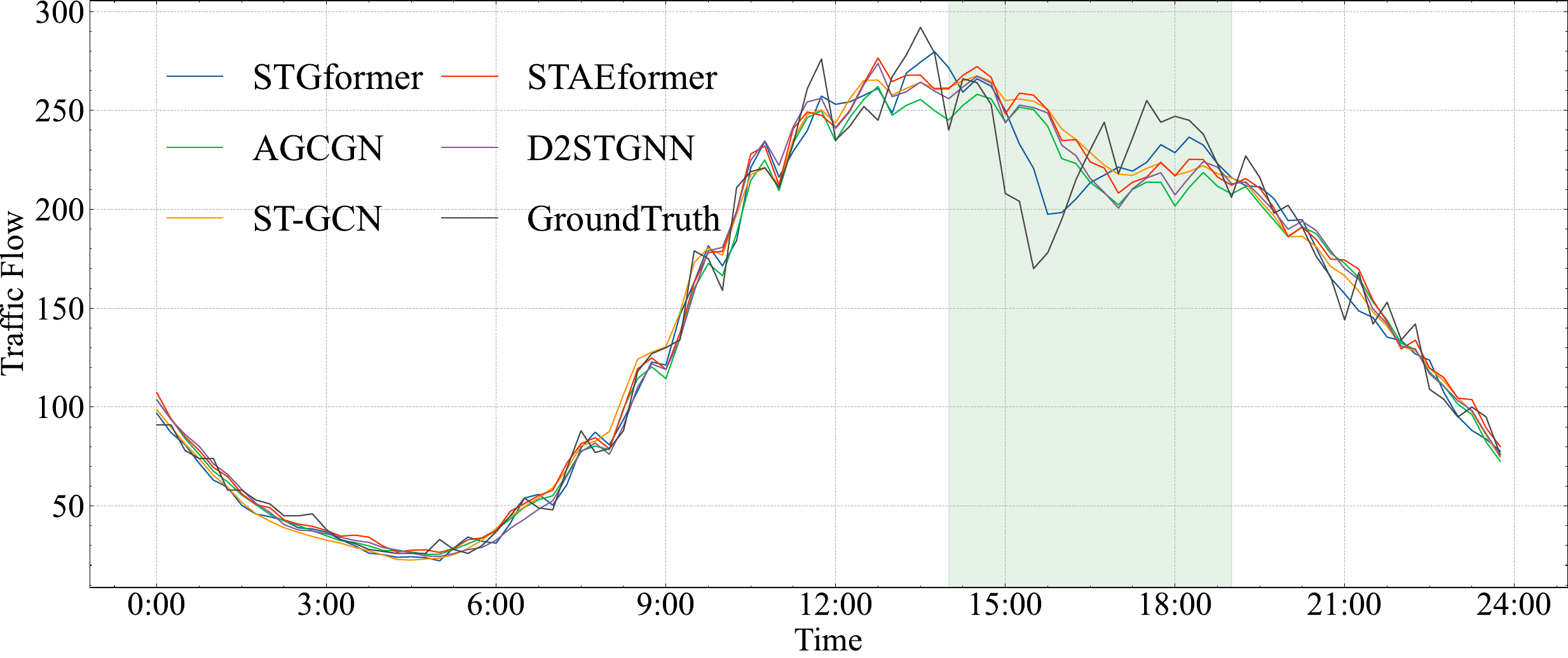}}
	\subfigure[Node 208 in  Los Angeles]{\includegraphics[width=0.48\linewidth]{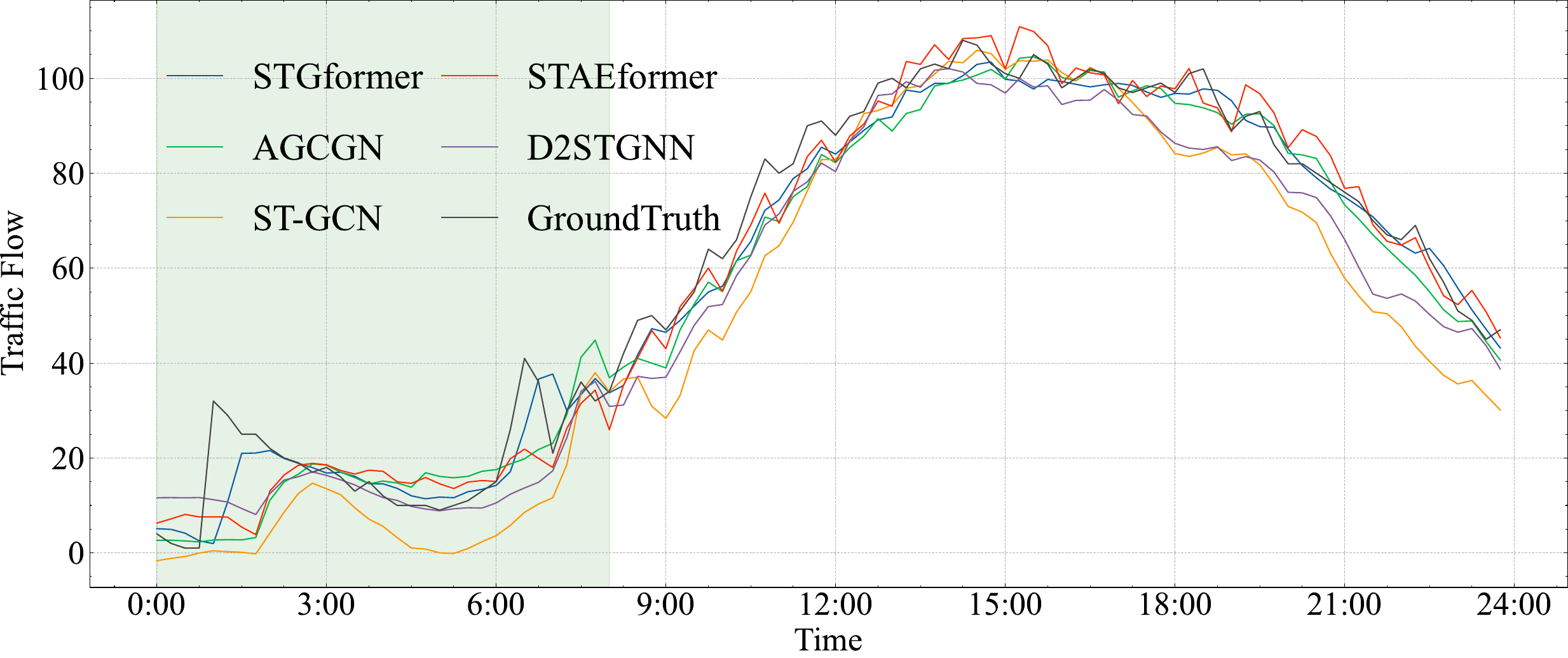}}
	
	\caption{\textbf{Model prediction analysis in one day on San Diego and Los Angeles datasets.}}
	\label{fig:case}
\end{figure*}

\subsection{Performance Comparisons.}

\noindent\textbf{Performance on LargeST.}  
We further evaluated  the performance  of STGformer on the LargeST datasets.
Experimental results in \tableref{tab:performance} demonstrate that STGformer consistently outperforms STAEformer across all evaluated datasets, exhibiting significant performance improvements. Specifically, in the San Diego dataset, STGformer achieved the most remarkable advancements in average metrics, with improvements of 3.61\%, 2.83\%, and 6.73\% in MAE, RMSE, and MAPE, respectively. While the improvement margins were relatively smaller for the Bay Area and Los Angeles datasets, STGformer maintained a consistent advantage, with average metric improvements ranging from 1.92\% to 2.12\%. Notably, STGformer not only excels in prediction accuracy but also demonstrates superior model efficiency. Taking the Los Angeles dataset as an example, STGformer achieved performance superior to STAEformer, which has 4.7M parameters, while utilizing only 705K parameters, highlighting its significant advantage in parameter efficiency. These results collectively indicate that STGformer can effectively enhance the accuracy of spatiotemporal sequence forecasting while maintaining a lower computational complexity, providing a valuable new direction for research in this field.

\noindent\textbf{Performance on Cross Year Scenario.}
\tableref{tab:ood} presents a comparative analysis of STGformer against other spatiotemporal baseline models in cross-year scenarios, including three major urban regions: San Diego, Bay Area, and Los Angeles. We compared the performance of spatiotemporal models trained on 2019 data when applied to 2020 data to evaluate model generalization ability. In the San Diego dataset, STGformer achieved a significant 14.14\% reduction in RMSE for 3-hour predictions, decreasing from 31.55 to 27.09. Similarly, on the Bay Area dataset, RMSE for the same forecast horizon decreased from 32.66 to 28.20, representing a substantial 13.66\% improvement. The Los Angeles dataset exhibited a similar trend, with a notable 13.10\% reduction in RMSE for 3-hour predictions, from 33.97 to 29.52. These results demonstrate that employing the STG attention block, as opposed to separate stacked attention mechanisms, offers greater robustness and adaptability when handling diverse urban traffic patterns and cross-year data variations, providing strong support for enhancing the accuracy of urban traffic flow forecasting.

\noindent\textbf{Performance on PEMS-series Benchmark.}
We validated the effectiveness of STGformer on the conventional PEMS-series datasets. As demonstrated in \tableref{tab:perf_pems}, the proposed STGformer model consistently outperforms STAEformer across all four PEMS datasets, showcasing its superior performance in traffic flow forecasting tasks. Notably, STGformer achieves the most significant improvement on the PEMS03 dataset, with a remarkable 8.97\% reduction in the RMSE metric (from 27.55 to 25.08). This result not only highlights STGformer's advantages in handling complex spatiotemporal data but also indicates its notable effectiveness in reducing prediction errors and enhancing model stability. Furthermore, STGformer achieves consistent performance improvements across all datasets while incurring only 0.2\% of the computational cost of STAEformer, further validating its adaptability and robustness in various traffic network environments.

\subsection{Ablation Study}\label{sec:Ablation}
To comprehensively evaluate the significance of various components within the STGformer, we conducted an ablation study using the LargeST dataset.  Four scenarios were considered in \figref{fig:abla2}: without self-attention (W/o SA), without temporal self-attention (W/o TSA), without spatiotemporal self-attention (W/o SSA), and without graph high-order interaction (W/o Graph).
Our findings indicate that the absence of any component led to a substantial decline in performance. Notably, the removal of spatiotemporal self-attention (W/o SA) resulted in the most significant performance degradation across all metrics (MAE, RMSE, and MAPE) and datasets, as it reduced the model to a feedforward module. Higher-order interactions (W/o Graph) also played a crucial role, with their absence having a relatively larger impact compared to removing only spatial or temporal self-attention.
Furthermore, spatial self-attention (W/o SSA) and temporal self-attention (W/o TSA) contributed significantly to performance, as their removal led to substantial performance drops. These results underscore the critical importance of each component, particularly global spatiotemporal information and higher-order interactions, for achieving optimal performance with the STGformer.

\subsection{Case Study} \figref{fig:case} illustrates the results of traffic flow prediction, including STGformer, STAEformer, AGCRN, D2STGNN, and ST-GCN, comparing the performance of multiple models across different nodes.
(a) Node 24 in San Diego: The STGformer model closely adheres to the ground truth traffic flow, demonstrating superior accuracy particularly during transitional periods between peak and off-peak hours. In contrast, other models exhibit deviations from the actual values at certain intervals.
(b) Node 64 in San Diego: STGformer's predictive curve aligns closely with the ground truth, notably outperforming other models in the afternoon period. Alternative models such as STAEformer and D2STGNN display greater fluctuations or discrepancies.
(c) Node 93 in Los Angeles: STGformer's performance is particularly noteworthy during morning and evening peak hours, with predictions nearly coinciding with the ground truth. Other models show varying degrees of overestimation or underestimation during these critical periods.
(d) Node 208 in Los Angeles: Despite all models exhibiting some bias in predictions during the afternoon traffic peak, STGformer demonstrates the closest approximation to actual flow trends. This is especially evident in the post-18:00 timeframe, where it significantly outperforms other models.

\section{Conclusion and Limitation}
In conclusion, this study introduces STGformer, a novel spatiotemporal graph Transformer model that addresses the computational challenges faced by existing GCN and Transformer-based methods in adapting to real-world road networks. STGformer demonstrates superior performance across various traffic benchmarks, from small-scale PEMS datasets to the large-scale LargeST dataset, utilizing only a single layer and linear spatiotemporal global attention. ST-Graph attention block enables efficient high-order spatiotemporal interactions for both global and local patterns, significantly reducing computational costs and memory usage compared to state-of-the-art methods like STAEformer. Furthermore, STGformer exhibits remarkable generalization capabilities, maintaining robust performance even when tested on data from a year later. These results position STGformer as a promising backbone for spatiotemporal models in large-scale traffic forecasting applications.




%
\bibliographystyle{IEEEtran}
\bibliography{reference}
\input{people}

\end{document}

%% file: people.tex
\begin{IEEEbiography}[{\includegraphics[width=1in,height=1.25in,clip,keepaspectratio]{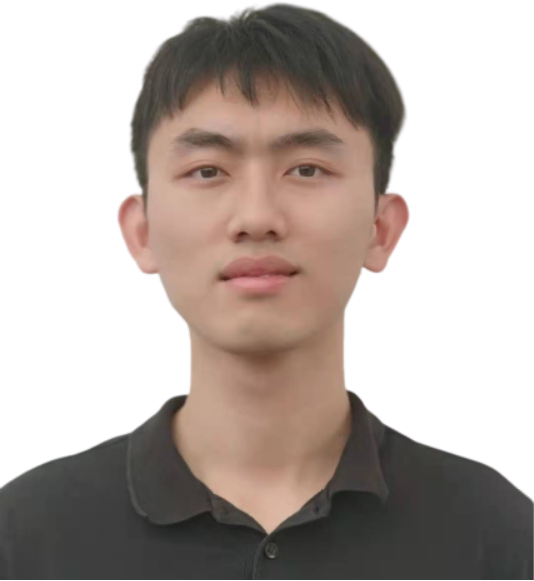}}]{Hongjun Wang} is working toward the PhD degree in the Department of Mechano-Informatics at The University of Tokyo. He received his M.S. degree in computer science and technology from Southern University of Science and Technology, China. He received his B.E. degree from the Nanjing University of Posts and Telecommunications, China, in 2019. His research interests are broadly in machine learning, with a focus on urban computing, explainable AI, data mining, and data visualization.
\end{IEEEbiography}
\vspace{1ex}
\begin{IEEEbiography}[{\includegraphics[width=1in,height=1.25in,clip,keepaspectratio]{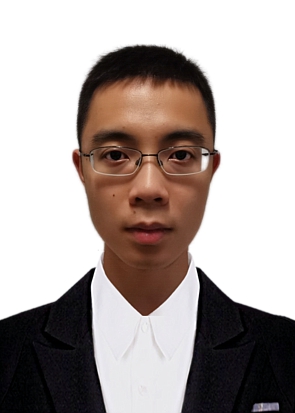}}]{Jiyuan Chen} is working towards his PhD degree at The Hong Kong Polytechnic University. He received his B.S. degree in Computer Science and Technology from Southern University of Science and Technology, China. His major research fields include artificial intelligence, deep learning, urban computing, and data mining.
\end{IEEEbiography}
\vspace{1ex}
\begin{IEEEbiography}[{\includegraphics[width=1in,height=1.25in,clip,keepaspectratio]{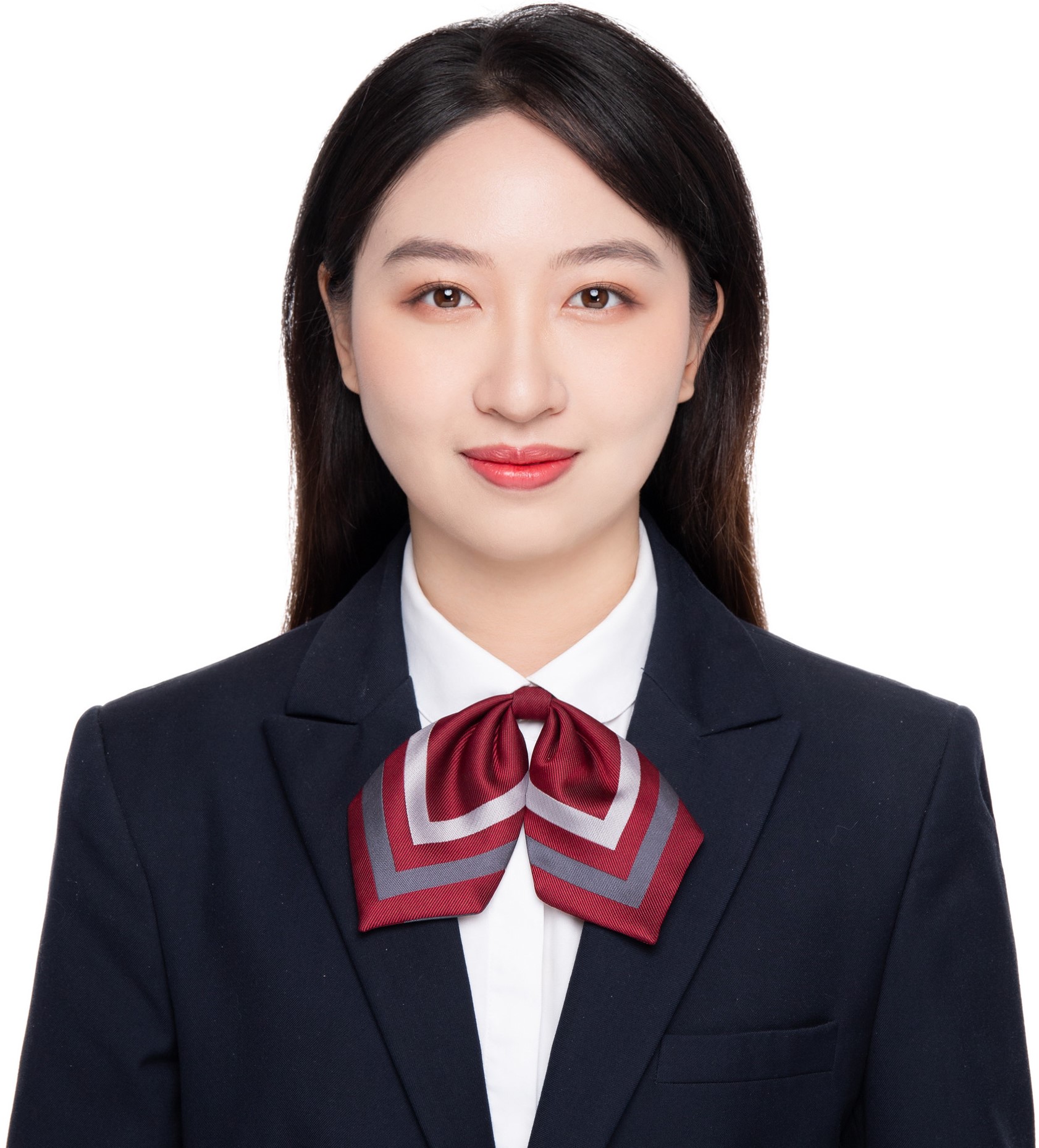}}]{Tong Pan} received a B.S. degree in Physics from East China Normal University, China, in 2019, and a Ph.D. degree in Physics from the Chinese University of Hong Kong, China, in 2024. From 2024, she has been a postdoctal researcher at Southern University of Science and Technology, China. Her research interests include data analysis, machine learning and AI for science. 
\end{IEEEbiography}
\vspace{1ex}
\begin{IEEEbiography}[{\includegraphics[width=1in,height=1.25in,clip,keepaspectratio]{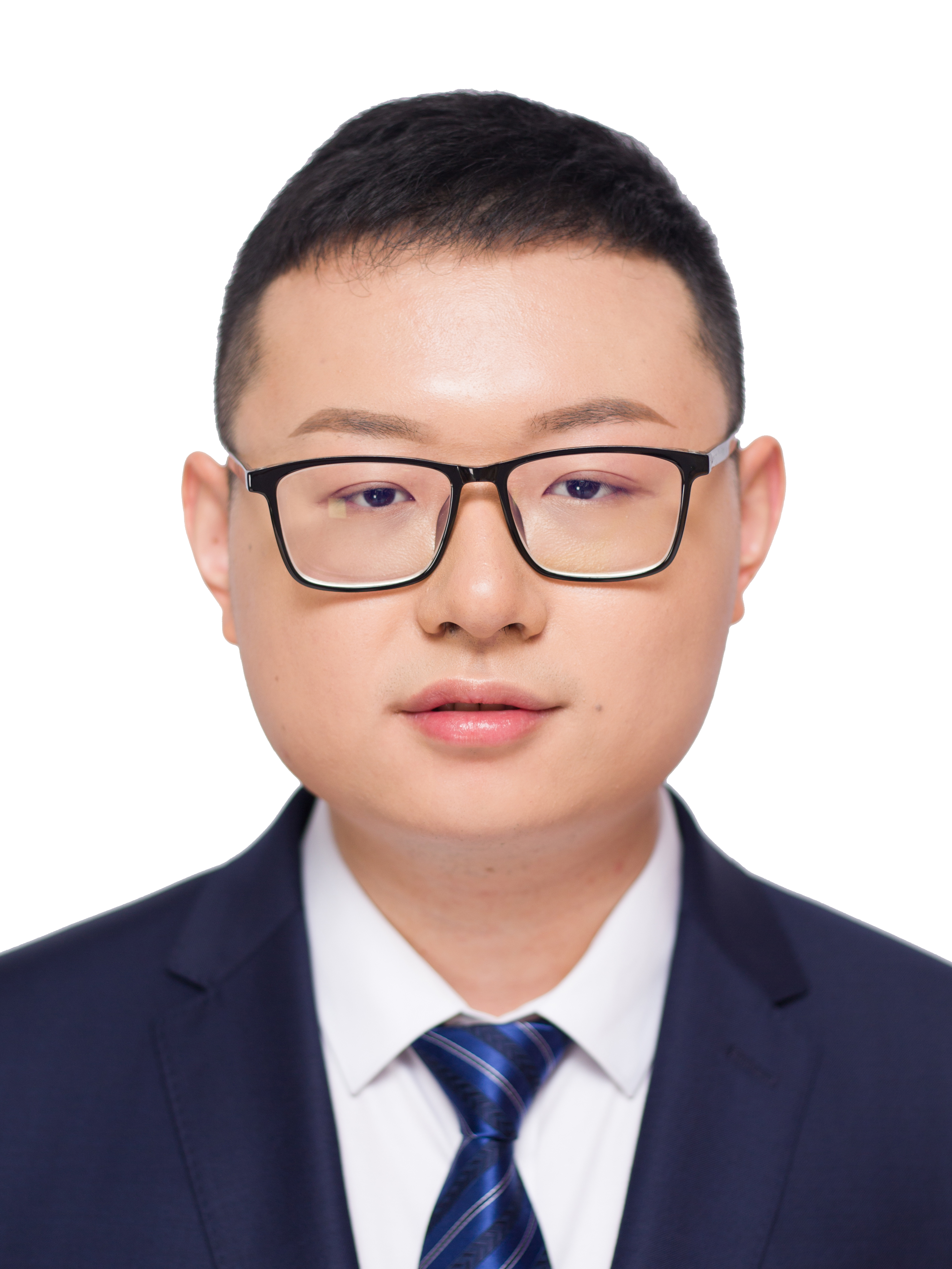}}]{Zheng Dong}  received his B.E. degree in computer science and technology from Southern University of Science and Technology (SUSTech) in 2022. He is currently persuing a M.S. degree in the Department of Computer Science and Engineering, SUSTech. His research interests include deep learning and spatio-temporal data mining.
\end{IEEEbiography}

\vspace{1ex}
\begin{IEEEbiography}[{\includegraphics[width=1in,height=1.25in,clip,keepaspectratio]{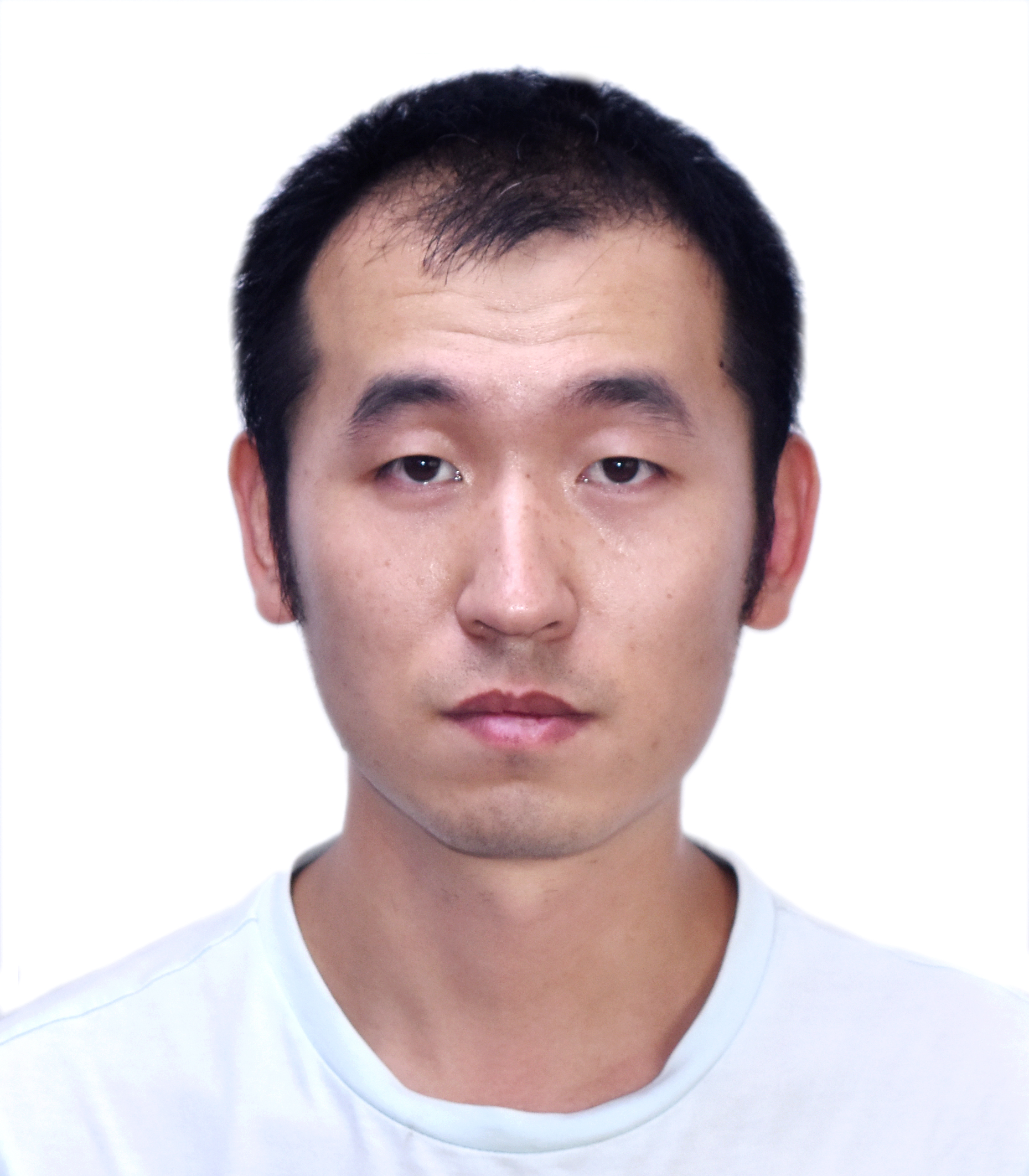}}]{Lingyu Zhang}   joined Baidu in 2012 as a search strategy algorithm research and development engineer. He joined Didi in 2013 and served as senior algorithm engineer, technical director of taxi strategy algorithm direction, and technical expert of strategy model department. Currently a researcher at Didi AI Labs, he used machine learning and big data technology to design and lead the implementation of multiple company-level intelligent system engines during his work at Didi, such as the order distribution system based on combination optimization, and the capacity based on density clustering and global optimization. Scheduling engine, traffic guidance and personalized recommendation engine, "Guess where you are going" personalized destination recommendation system, etc. Participated in the company's dozens of international and domestic core technology innovation patent research and development, application, good at using mathematical modeling, business model abstraction, machine learning, etc. to solve practical business problems. He has won honorary titles such as Beijing Invention and Innovation Patent Gold Award and QCon Star Lecturer, and his research results have been included in top international conferences related to artificial intelligence and data mining such as KDD, SIGIR, AAAI, and CIKM.
\end{IEEEbiography}
\vspace{1ex}
\begin{IEEEbiography}[{\includegraphics[width=1in,height=1.25in,clip,keepaspectratio]{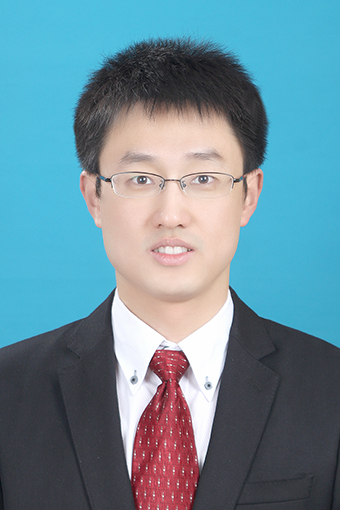}}]{Renhe Jiang} received a B.S. degree in software engineering from the Dalian University of Technology, China, in 2012, a M.S. degree in information science from Nagoya University, Japan, in 2015, and a Ph.D. degree in civil engineering from The University of Tokyo, Japan, in 2019. From 2019, he has been an Assistant Professor at the Information Technology Center, The University of Tokyo. His research interests include ubiquitous computing, deep learning, and spatio-temporal data analysis.
\end{IEEEbiography}
\vspace{1ex}
\begin{IEEEbiography}[{\includegraphics[width=1in,height=1.25in,clip,keepaspectratio]{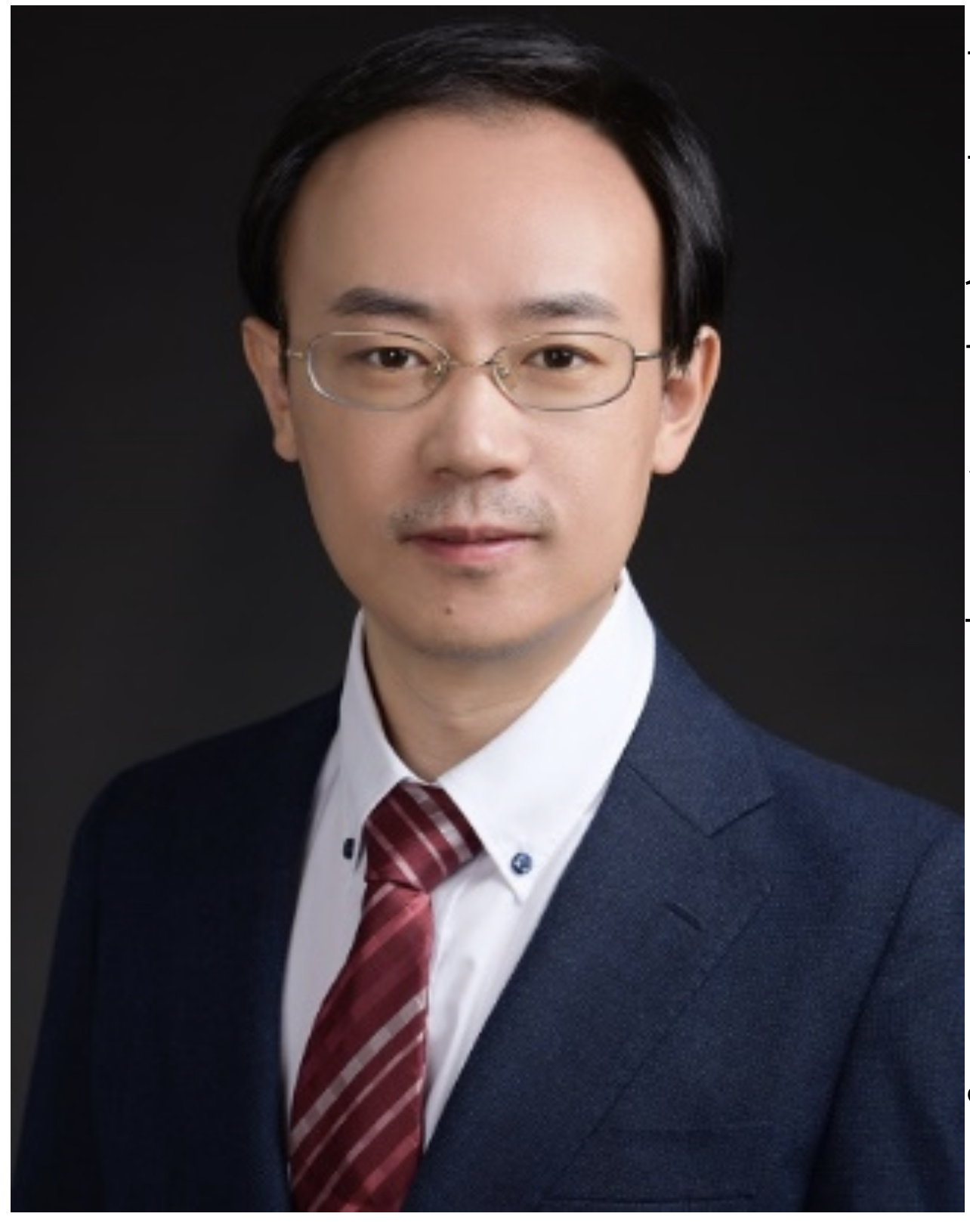}}]{Prof. Xuan Song}  received the Ph.D. degree in signal and information processing from Peking University in 2010. In 2017, he was selected as an Excellent Young Researcher of Japan MEXT. In the past ten years, he led and participated in many important projects as a principal investigator or primary actor in Japan, such as the DIAS/GRENE Grant of MEXT, Japan; Japan/US Big Data and Disaster Project of JST, Japan; Young Scientists Grant and Scientific Research Grant of MEXT, Japan; Research Grant of MLIT, Japan; CORE Project of Microsoft; Grant of JR EAST Company and Hitachi Company, Japan. He served as Associate Editor, Guest Editor, Area Chair, Program Committee Member or reviewer for many famous journals and top-tier conferences, such as IMWUT, IEEE Transactions on Multimedia, WWW Journal, Big Data Journal, ISTC, MIPR, ACM TIST, IEEE TKDE, UbiComp, ICCV, CVPR, ICRA and etc.
\end{IEEEbiography}